\documentclass[a4paper]{article}
\usepackage[margin=1in]{geometry}
\usepackage{microtype}
\usepackage{graphicx}
\usepackage{subfigure}
\usepackage{booktabs} 
\usepackage{natbib}

\newcommand{\N}{\mathcal{N}}
\newcommand{\rx}{\overset{\rightarrow}{X}}
\newcommand{\lx}{\overset{\leftarrow}{X}}
\newcommand{\ly}{\overset{\leftarrow}{Y}}
\newcommand{\e}{\mathbb{E}}
\newcommand{\lp}{\overset{\leftarrow}{P}}
\newcommand{\rp}{\overset{\rightarrow}{P}}
\newcommand{\Tr}{\mathrm{Tr}}
\newcommand{\q}{\overset{\leftarrow}{Q}}
\newcommand{\rxdt}{\overset{\rightarrow}{X^D_t}}
\newcommand{\lxdt}{\overset{\leftarrow}{X^D_t}}
\newcommand{\lxnt}{\overset{\leftarrow}{X^N_t}}
\newcommand{\lxdT}{\overset{\leftarrow}{X^D_T}}
\newcommand{\lxnT}{\overset{\leftarrow}{X^N_T}}

\newcommand{\rpdt}{\overset{\rightarrow}{P^D_t}}

\newcommand{\rpdo}{\overset{\rightarrow}{P^D_0}}

\newcommand{\R}{\mathbb{R}}

\usepackage{amsmath,amssymb,amsthm}
\usepackage{hyperref}
\usepackage[shortlabels]{enumitem}
\setlist{nolistsep}
\usepackage[capitalize,noabbrev]{cleveref}
\theoremstyle{plain}
\newtheorem{assumption}{Assumption}
\newtheorem{theorem}{Theorem}[section]
\newtheorem{proposition}[theorem]{Proposition}
\newtheorem{lemma}[theorem]{Lemma}
\newtheorem{corollary}[theorem]{Corollary}
\theoremstyle{definition}

\theoremstyle{remark}
\newtheorem{remark}[theorem]{Remark}

\usepackage{xcolor}

\DeclareMathOperator{\diam}{Diam}
\DeclareMathOperator{\supp}{Supp}
\newcommand{\A}{{\bar A}}

\title{Wasserstein Bounds for generative diffusion models with Gaussian tail targets}

\author{Xixian Wang\thanks{Division of Mathematical Sciences, School of Physical and Mathematical Sciences, Nanyang Technological University, Singapore. xixian001@e.ntu.edu.sg}, Zhongjian Wang\thanks{Division of Mathematical Sciences, School of Physical and Mathematical Sciences, Nanyang Technological University, Singapore. zhongjian.wang@ntu.edu.sg (Corresponding)}}

\begin{document}

\maketitle

\begin{abstract}
We present an estimate of the Wasserstein distance between the data distribution and the generation of score-based generative models. The sampling complexity with respect to dimension is $\mathcal{O}(\sqrt{d})$, with a logarithmic constant. In the analysis, we assume a Gaussian-type tail behavior of the data distribution and an $\epsilon$-accurate approximation of the score. Such a Gaussian tail assumption is general, as it accommodates a practical target - the distribution from early stopping techniques with bounded support.

The crux of the analysis lies in the global Lipschitz bound of the score, which is shown from the Gaussian tail assumption by a dimension-independent estimate of the heat kernel. Consequently, our complexity bound scales linearly (up to a logarithmic constant) with the square root of the trace of the covariance operator, which relates to the invariant distribution of the forward process.
\end{abstract}


\section{Introduction}
\label{sec:intro}

Diffusion models (DM) are among the state-of-the-art tools in the new GenAI era. As generative models, diffusion models first links the target distribution to some distribution easy to sample via a diffusive process (forward). The generative (backward) process then \emph{reverses} the diffusion, enabling samples from the easily sampled distribution to be transformed into samples from the target distribution. A well-known mathematical model that encapsulates this approach is the score-based stochastic differential equation (SDE) model \citep{song2020score}, where the forward and backward processes are represented by two SDEs that share the same marginal distribution \citep{And_80,haussmann1986time}. In most cases, the forward process is assumed to be an Ornstein-Uhlenbeck (OU) process. The backward process incorporates the gradient of the logarithmic density (score) of the forward process. When the explicit form of the score is unknown, it is estimated by a neural network from discrete samples of the target distribution. 

A major direction for the theoretical study of DMs is the convergence of the approximated backward process with limited data assumptions. When the Lipschitz bound of the score is available (regular target), the backward process is generally well-defined until $t=0$, and convergence results are related to the Lipschitz bound. Otherwise, when the bound is unavailable (singular target), the early stopping technique is introduced, and convergence results are related to the stopping time. Various analytical approaches have been adapted for these two types of assumptions.

\textbf{In this work}, we aim to present a general error analysis that applies to both regular and singular target distributions and also generalizes to an infinite-dimensional setting. The analysis is based on \emph{a dimensionless and global-in-space Lipschitz bound} of the score, derived from a heat kernel estimation approach. Our complexity bounds, derived under the \emph{Wasserstein-2-metric}, are \emph{linear (with a logarithmic constant) in the square root of variance} of the Brownian motion in the forward process (Theorem~\ref{thm:w2 bound}). In the finite-dimensional case with the standard Gaussian as the base distribution (the invariant measure of forward process), the variance is linear in the dimension and hence the complexity is $\mathcal{O}(\sqrt{d})$. For the infinite-dimensional case, a Gaussian random field is taken as the base distribution, and the variance is linear with the trace of the covariance operator.
\begin{table}[t]
\caption{Summary of previous bounds and our results for score-based diffusion models in $d$ dimensions. The complexity bound gives the number of steps $N$ needed to ensure error $\leq \epsilon_0$.}
\label{tab:complexity}
\vspace{-0.5em}
\begin{center}
\scriptsize

\begin{tabular}{@{}p{3.0cm}p{2.0cm}p{4.4cm}p{5.2cm}@{}}
\toprule
Target $\rp_0$ & Metric & Complexity & Result \\
\midrule
Supp $B_R(0)$ & $\mathcal{W}_1(\rp_0, \q_{T-\delta})$ & 
\shortstack[l]{$\log N = \mathcal{O}\Bigl(\frac{R^2(d+R^4)^2(\log R)^2}{\epsilon_0^2}\Bigr)$} &
\citep{de_bortoli_convergence_2023} Thm.~1 + Cor.~2 \\

Supp $B_R(0)$ & $\mathcal{W}_2(\rp_0, \q_{T-\delta})$ & 
\shortstack[l]{$\log N = \mathcal{O}\Bigl(\frac{R^2 d^2}{\epsilon_0^4}\Bigr)$} &
This work: Cor.~\ref{cor:complexity under manifold ass 3} \\

$\e|X_0|^2<\infty$ & KL$(\rp_\delta || \q_{T-\delta})$ & 
\shortstack[l]{$N = \mathcal{O}\Bigl(\frac{d \log^2 \tfrac{1}{\delta}}{\epsilon_0}\Bigr)$} &
\citep{benton2023linear} Cor.~1 \\
$\e|X_0|^2<\infty$ & TV$(\rp_0 || \q_{T})$ & 
\shortstack[l]{$N = \mathcal{O}\Bigl(\frac{d }{\epsilon_0}\Bigr)$} &
\citep{li2025odtconvergencetheorydiffusion} Thm.~1\\

$\nabla \log p_0$ $L$-lip & KL$(\rp_0 || \q_T)$ & 
\shortstack[l]{$N = \mathcal{O}\Bigl(\frac{d^2}{\epsilon_0}\Bigr)$} &
\citep{chen2023improvedanalysisscorebasedgenerative} Thm.~5 \\

$\mathcal{F}(\rp_0|\N(0,I_d)) \lesssim d\ ^{(*)}$   & KL$(\rp_0 || \q_T)$ & 
\shortstack[l]{$N = \mathcal{O}\Bigl(\frac{d}{\epsilon_0^2} \log \frac{d}{\epsilon_0}\Bigr)$} &
\citep{conforti2024klconvergenceguaranteesscore} Thm.~1 \\

$P_0$ log-concave & $\mathcal{W}_2(\rp_0, \q_T)$ & 
\shortstack[l]{$N = \mathcal{O}\Bigl(\frac{d}{\epsilon_0^2} \log \frac{d}{\epsilon_0}\Bigr)$} &
\citep{gao2023wassersteinconvergenceguaranteesgeneral} Tab.~2 \\

$P_0$ one-side Lipschitz + weakly log-concave & $\mathcal{W}_2(\rp_0, \q_T)$ & 
\shortstack[l]{$N = \mathcal{O}\Bigl(\frac{d^2}{\epsilon_0^2} \Bigr)$} &
\citep{silveribeyond} Thm~3.5 \\

G tail, Ass.~\ref{ass:Gtail} & $\mathcal{W}_2(\rp_0, \q_T)$ & 
\shortstack[l]{$N = \mathcal{O}\Bigl(\frac{\sqrt{d}}{\epsilon_0} \log \frac{d}{\epsilon_0^2}\Bigr)$} &
This work: Cor.~\ref{cor:complexity} \\
\bottomrule
\end{tabular}

\end{center}
\vspace{-1em}
\begin{flushleft}
\scriptsize
* $\mathcal{F}$ denotes relative Fisher information. Our Gaussian tail assumption (Assumption~\ref{ass:Gtail}) implies $\mathcal{F}(\rp_0|\N(0,I_d)) \lesssim d$, and they are equivalent under the standard Gaussian case.
\end{flushleft}
\vspace{-1.5em}
\end{table}
It is worth noting that the results of this work adopt the \emph{Wasserstein-$2$ distance} as the metric of error (instead of Kullback–Leibler (KL) divergence) due to its flexibility. More precisely, the reasons are two-fold: \textbf{(1)} In practical applications of diffusion models for the generation of structured data (image, text, video, protein, etc.), the target distributions mostly find their support in a compact sub-manifold, see further discussion of the manifold hypothesis in \citep{tenenbaum2000global,bengio2013representation,bengio2017deep}. Consequently, under this hypothesis, the standard KL divergence cannot be consistently defined between the distribution obtained via the backward process, whose support is the entire space, and the target distribution with compact support. \textbf{(2)} In high (towards infinite) dimension settings, it becomes necessary to compactify the forward process. One way to achieve this is by choosing the covariance matrix (operator) $C$ in the forward process~\eqref{eq:forward process} to have finite trace, so that the invariant distribution of the process has a finite second moment. The Wasserstein distance then scales with $\Tr(C)$, making it consistent with this compactification in the context of infinite-dimensional generative models. By contrast, the KL divergence scales with the ambient dimension and therefore cannot be directly extended to yield a dimensionless result. A motivating example is provided in Appendix~\ref{app:klvsw}.

\paragraph{Related work} ~~

\underline{Complexity bounds}
\citep{deconvergence}
established the first convergence guarantees in the 1-Wasserstein distance, assuming that the
data distribution satisfies the so-called manifold hypothesis. Some recent works 
\citep{pierret2024diffusionmodelsgaussiandistributions,gao2024convergenceanalysisgeneralprobability,gao2023wassersteinconvergenceguaranteesgeneral,silveribeyond,li2025odtconvergencetheorydiffusion,bruno2025wasserstein} 
established convergence guarantees under Gaussian, log-concave, weakly log-concave, or other minimal assumptions. 
We are also aware of several complexity bounds under KL divergence, for instance \citep{chen2023improvedanalysisscorebasedgenerative,benton2023linear,conforti2024klconvergenceguaranteesscore}. A common point of these approaches is the use of the chain rule for KL divergence, which follows from the  Girsanov theorem, to separate the global error into local truncation errors. To bound local ones, the probabilistic viewpoint, which estimates the score function under the distribution of the forward process, comes into play. In this work, we instead 
provide a point-wise estimate of the score. Such an estimate also facilitates the analysis of the score under the approximated backward process, thereby providing a Wasserstein bound.
In Table~\ref{tab:complexity}, we present the comparison of the complexity bounds.

\underline{Diffusion models in infinite dimension}
Typically, diffusion models operate on 
finite-dimen-sional spaces. However, in many domains, the underlying signal is infinite-dimensional, and the observed data is a collection of discrete observations of some underlying function, for instance,  Bayesian
inverse problems \citep{stuart2010inverse}.
Up to now, there are many studies applying diffusion models to functional spaces \citep{lim_score-based_2023,kerrigan_diffusion_2023,pidstrigach_infinite-dimensional_2023,franzese2024continuous}. 
While many theoretical studies \citep{de_bortoli_convergence_2023,chen2023improvedanalysisscorebasedgenerative,benton2023linear} suggest that performance guarantees deteriorate
 with increasing dimension. \citep{pidstrigach_infinite-dimensional_2023} established a bound on the Wasserstein-2 Distance from the samples to the target distribution, which is dimension-independent but grows exponentially with the running time $T$.
 In this work, we further provide a Wasserstein-2-bound that is both dimension-independent and running time uniform under the Gaussian tail assumption.

\underline{Lipschitz bounds of the score} The score functions in the SGMs are related to the gradient of log-density ($\log p$) of the forward process. It is well known that the function $\log p$ itself follows a viscous Hamilton-Jacobi (vHJ) equation, as seen in~\eqref{PDE1} in the later discussion. Then the Lipschitz bounds of the scores are equivalent to Hessian bounds for a vHJ equation. There are various regularity results in the literature for the original Fokker-Planck equation or the transformed vHJ, see
\citep{fujita2006asymptotic,stromberg2010semiconcavity} and recent results in \citep{blessing_viscous_2022,mooney2024global}. We would point out that, except \citep{mooney2024global}, most results are seeking a spatially global Hessian bound which only lasts for a finite time without the Gaussian tail assumption (Assumption~\ref{ass:Gtail}) in this work.  \citep{mooney2024global} also provides a local-in-space and global-in-time bound, while only polynomial in dimension ($d^3$) complexity can be shown from it due to the spatial locality. 

We are also aware of the literature on the topic of contraction properties or Lipschitz estimates of transport maps between measures.\footnote{More precisely, in the context of this work, the transport map refers to the push-forward map from the Gaussian (base distribution) to the target distribution $p_0$.} The \emph{Caffarelli's contraction theorem} \citep{caffarelli2000monotonicity} in the optimal transport (OT) setup is the starting point, and see \citep{colombo2015lipschitz} for generalization to Lipschitz estimate. Besides OT, the contractive map can also be attained by (reverse) heat flow \citep{kim_generalization_2012}. Recent generalizations to the Lipschitz estimates of the maps, including \citep{mikulincer2023lipschitz,neeman2022lipschitz,fathi2024transportation,brigati2024heat}, obtain results for the boundedness of flow in a similar fashion to this work under diverse assumptions. In comparison, this work focuses more on deriving theories with applications to score-based diffusion models. In this regard, our assumptions cover the spatially anisotropic noise (no necessity for equivalence among $A$, $C$, and $I_d$ in Assumption~\ref{ass:Gtail}) and we {discuss} the convergence and complexity bounds in the discrete approximation of the generative flows.

\textbf{The main contributions of this paper are:}
\begin{itemize}
    \item  We introduce a Gaussian tail assumption (Assumption~\ref{ass:Gtail}) and apply a heat kernel estimate (Theorem~\ref{thm:vHJ hessian bound}) inspired by the solution of the viscous Hamilton–Jacobi (vHJ) equation to obtain a spatially global Lipschitz constant for the modified score function that decays exponentially over time (Corollary~\ref{cor:score lip exp decay bound}). This assumption accommodates non-log-concave target distributions and covers practical scenarios, such as when the support of the target distribution is bounded and the early stopping technique \citep{lyu2022accelerating} is employed (Theorem~\ref{thm:MH}).
    \item We establish a Wasserstein-2 bound (Theorem\ref{thm:w2 bound}) that depends only on the second moment of the base distribution, allowing the result to extend naturally to infinite-dimensional settings. When the second moment scales proportionally with the ambient dimension, the resulting iteration complexity achieves the state-of-the-art dependence on the dimension, with a rate of $\mathcal{O}(\sqrt{d})$. \citep{gao2023wassersteinconvergenceguaranteesgeneral} Proposition 8 also shows that under the standard Gaussian distribution, such a complexity bound is optimal.
    \item As part of our methodological contribution, we provide a modified approach rather than conventional Lyapunov-type analysis, which utilizes the exponential decay structure of the score to show the uniform boundedness of accumulation error for an arbitrary long forward/backward process.
\end{itemize}
\section{Preliminaries}

\subsection{Background and Setting the Stage}

A large class of generative diffusion models can be analyzed under the SDE framework \citep{song2020score}, which contains two processes: forward and backward. The forward process, which gradually transforms the data distribution into white noise, is an OU process as follows,
\begin{equation}
     d\rx_t= -\frac{1}{2}\rx_tdt+\sqrt{C}dB_t, \quad   0\le t\le T.
     \label{eq:forward process}
\end{equation}
where $B_t$ is a standard Brownian motion, $C$ is a symmetric, positive-definite covariance matrix and $T$ is the final time such that the distribution of $X_T$ is close to the Gaussian distribution $\N(0,C)$, denoted as \emph{base distribution}. The initial $\rx_0$ follows the target(data) distribution, denoted as $\rp_0$. We remark that in the majority of theories and applications of the diffusion model, $C$ is assumed to be the identity. While we keep the spatially anisotropic noise assumption ($C\not\equiv I_d$) in the following derivation to enable our theories to generalize to an infinite-dimensional setting where some compactification is necessary, see also \citep{pidstrigach_infinite-dimensional_2023}. We denote the probability density of the forward process $\rx_t$ by $p_t$,
then $p_t$ solves the Fokker-Planck equation with Cauchy data $p_0$:
\begin{equation}
    \begin{cases}
        \partial_tp=\frac{1}{2}\big(\nabla\cdot(xp)+\nabla \cdot C\nabla p\big),\\
        p(0,x)=p_0(x).
    \end{cases}
    \label{eq:fokker planck}
\end{equation}
With time reversal $\lx_t:=\rx_{T-t}$, the backward process $(\lx_t)_{0\le t\le T}$ satisfies the following SDE~\citep{haussmann1986time},
\begin{equation}
    d\lx_t= \big(\frac{1}{2}\lx_t+s(T-t,\lx_t)\big)dt+\sqrt{C}d\tilde{B}_t,
    \label{eq:backward process}
\end{equation}
where $\tilde{B}_t$ is also a standard Brownian motion (may not be the same as $B_t$) and the term $s(t,x):=C\nabla \log p_t(x)$ is generally referred to as the score function. The process $(\lx_t)_{0 \le t \le T}$ transforms noise into samples follows $\rp_0$.
We denote $\rp_t$(correspondingly $\lp_t$) as the marginal distribution of $\rx_t$ in~\eqref{eq:forward process}($\lx_t$ in~\eqref{eq:backward process}). Then, $\forall 0 \le t \le T$, $\lp_t=\rp_{T-t}$, especially $\lp_T=\rp_0$.

In practice, the score function $s(t,x)=C\nabla \log p_{t}(x)$ is not available since the closed form expression of $p_0$ is unknown. Thus, we model the score function by a neural network $s_\theta(t,x)$, where $\theta$ denotes latent variables of the neural network. We train the network by optimizing an $L_2$ estimation loss, \begin{equation}
    \nonumber
    \e_{\rp_t}\|s_\theta(t,x)-C\nabla \log p_t(x)\|^2.
\end{equation}
Given the estimated score $s_\theta$ (assumed to be $\epsilon$-accurate, specified in Assumption~\ref{ass:error}), one can generate samples of the target distribution $p_0$ by a numerical approximation of the backward process starting from the Gaussian distribution $\N(0,C)$,
\begin{equation}
    d\ly_t= \big(\frac{1}{2}\ly_t+s_\theta(T-t,\ly_{t})\big)dt+\sqrt{C}d\bar{B}_t.
\end{equation}
\paragraph{Time discretization}
 We employ an Euler-type odiscretization of the continuous-time stochastic process, which facilitates the convergence analysis. Let $0=t_0 \le t_1 \le \cdots \le t_N=T-\delta$ be the time discretization points (schedule), $\delta=0$ for the normal setting and $\delta > 0$ for the
early-stopping setting \citep{lyu2022accelerating}. In this work, we adopt the following discrete scheme. Starting from $\ly_0\sim\N(0,C)$, for all $ k =0,1,...,N-1,$
\begin{equation}\label{eq:explicit solver}
    \begin{aligned}
\ly_{t_{k+1}}
        =\frac{1}{\sqrt{\alpha_k}}\left(\ly_{t_k}+(1-\alpha_k)s_\theta(T-t_k,\ly_{t_k})\right)+\sqrt{1-\alpha_k}\bar{z}_k,
\end{aligned}
\end{equation}
where $\alpha_k=\exp(t_k-t_{k+1})$ and $\bar{z}_k\sim\N(0,C)$ are i.i.d.
\begin{remark}
    The discretization~\eqref{eq:explicit solver} approximately corresponds to the discrete-time scheme introduced in \citep{ho2020denoising}; it can also be found in the analysis work \citep{de_bortoli_convergence_2023}.  
\end{remark}

\paragraph{General Notations} Let $\gamma_C$ be the Gaussian measure $\N(0,C)$. For an $n \times n$ matrix $A$, we use the operator norm $\|\cdot\|$,
\begin{equation}
    \nonumber
    \|A\|=\sup_{v \neq0}\frac{|Av|}{|v|}:= \text{the \ largest \ eigenvalue \ of \ }\sqrt{A^TA}.
\end{equation}
For a symmetric, positive-definite $n \times n$ matrix $A$, we use $|\cdot|_A$ to denote weighted $l_2$ norm in $\mathbb{R}^n$ such that,\begin{equation}
    \nonumber 
    |x|_A^2:=\langle A^{-1/2}x ,A^{-1/2}x\rangle.  
\end{equation}
When $A$ is the identity matrix, we neglect the letter for simplicity and  $|\cdot|$ is the standard $l_2$ norm in $\mathbb{R}^n$.
For vector (matrix, correspondingly) valued function $f$ with $x$ as variable, $|f|_\infty:=\sup_x|f(x)|$ (correspondingly, $\|f\|_\infty:=\sup_x\|f(x)\|$).

For compactness of the process, we made the following assumption to bound the second moment of the forward process.
\begin{assumption}\label{ass:bound 2 moment}
    The data distribution has a bounded second moment, $
M_2:=\e_{\rp_0}|x|^2 < \infty$. The covariance $C$ in \eqref{eq:forward process} is in trace-class. 
    Furthermore, we denote, 
    \begin{align}
        M_0=\max\{\Tr(C),M_2,1\}\label{def:M0}.
    \end{align}
    
\end{assumption}

\subsection{Foundational Ideas based on Heat Kernel Estimation}
The convergence analysis of the discrete scheme~\eqref{eq:explicit solver} relies heavily on the regularity (up to second order) of the score potential function $\log p_t$, which follows a viscous Hamilton-Jacobi equation (vHJ). While similar derivations are known to experts \citep{evans2022partial}, we provide a brief list of derivations here for completeness. 

We first consider the following transform of $\log p_t$,
\begin{equation}\label{eq:score potential q}
    q(t,x)=-\log p_t(x)-\frac{x^T\A_t^{-1}x}{2},
\end{equation}
where $\A_t=Ae^{-t}+C(1-e^{-t})$. Then, $q$ satisfies the vHJ,
\begin{align} \label{PDE1}
    \left\{
    \begin{array}{l}
        \partial_t q-\frac{1}{2}\nabla\cdot C\nabla q +\frac{1}{2}|\sqrt{C}\nabla q|^2+(C\A^{-1}_t-\frac{1}{2}I_d)x\cdot\nabla q=\frac{1}{2}\Tr(C\A^{-1}_t-I_d),\\
    q(0,x)=-h(x) .
    \end{array}
    \right.
\end{align}
To simplify~\eqref{PDE1}, we let $f(t)=-\frac{1}{2} \int_0^t\Tr(C\A^{-1}_s-I_d)ds$ and make a two step change of variables:
let $K(t)=(A\A^{-1}_t)e^{-\frac{t}{2}}$, then 
    $\bar{q}(t,x)=q(t,K(t)^{-1}\sqrt{C}x)+f(t)$
satisfies, \begin{equation}
\begin{cases}
    \partial_t\bar{q}-\frac{1}{2}\nabla \cdot K(t)^2\nabla \bar{q}+\frac{1}{2}|K(t)\nabla \bar{q}|^2=0,\\
    \bar{q}(0,x)=\bar{h}(x):=-h(\sqrt{C}x).
    \label{PDE2}
\end{cases}    
\end{equation}
Lastly, we define $\bar{p}(t,x):=e^{-\bar{q}(t,x)}$
which satisfies \begin{equation}\label{eq:PDE3}
    \begin{cases}
        \partial_t\bar{p}-\frac{1}{2}\nabla\cdot K(t)^2\nabla\bar{p}=0, \quad \text{on} \ (0,\infty) \times \mathbb{R}^n,\\
        \bar{p}(0,x)=e^{-\bar{h}(x)}.
    \end{cases}
\end{equation}
\eqref{eq:PDE3} is a heat equation and admits the following solution from the heat kernel,
\begin{equation}\label{eq:solution of PDE3}
  \bar{p}(t,x)=  \frac{1}{(2\pi)^\frac{n}{2}}\int _{\mathbb{R}^n}\frac{1}{\sqrt{\det B(t)}}\exp\Big({\frac{-|x-y|_{B(t)}^2}{2}}\Big)\exp\big(-\bar{h}(y)\big)dy,
\end{equation}
where $B(t):=\int_0^tK(s)^2ds=(e^{\frac{t}{2}}-e^{-\frac{t}{2}})K(t)$. It is worth mentioning that~\eqref{eq:solution of PDE3} provides a solution to vHJ~\eqref{PDE1}, thus, analyzing~\eqref{eq:solution of PDE3} amounts to establishing the regularity of the score function $\log p_t(x)$.

\section{Results}
In this section, we list the theoretical results, and their detailed proofs are provided in Appendix~\ref{app:proofs}. Section~\ref{sec:Hbound} is devoted to the Lipschitz bound of score with Gaussian tail assumption, which is based on a heat kernel estimation of~\eqref{eq:solution of PDE3}. Section~\ref{sec:mainbound} lists the fundamental convergence result in the Wasserstein metric. Sections~\ref{sec:manifold} and~\ref{sec:beysian} present applications of our convergence result to the bounded-support assumption and Bayesian inverse problems, respectively. 
\subsection{Lipschitz Bound of Score  Function}\label{sec:Hbound}
The foundation of our analysis is to provide a spatially global regularity estimate of $p_t$ as the solution of an elliptic equation. Before that, we need the following assumption to derive reasonable point-wise estimates of the gradients of the score
function, 
\begin{assumption}[Gaussian tail]\label{ass:Gtail} The density of target distribution $p_0\in C^2(\R^d)$ and admits  the  tail decomposition,
\begin{equation}\label{tail_exp}
    p_0(x)= \exp \Big(-\frac{|x|_A^2}{2}\Big)\exp\big(h(x)\big),
\end{equation}
where 
\begin{enumerate}[(i),noitemsep]
    \item $A$ is a symmetric, positive-definite matrix which can be simultaneously diagonalized with $C$ and satisfies
   $\|AC^{-1}\|<\infty$ and $ \|CA^{-1}\|<\infty,$
  \item The remainder term $h$ satisfies $ |\sqrt{C}\nabla h|_\infty< \infty$ and $ \|C\nabla^2 h\|_\infty<\infty$.
\end{enumerate}
Moreover, all the constants above are dimension-independent.
\end{assumption}
\eqref{tail_exp} is equivalent to the following in log density form \begin{align*}
    \log p_0(x) = h(x) -\frac{|x|_A^2}{2}.
\end{align*}

    The Gaussian tail Assumption~\ref{ass:Gtail} is slightly more restrictive than both the $ L$-Lipschitz condition on $\nabla\log p_0$ and the combination of weak log-concavity with one-sided Lipschitz-ness in \citep{silveribeyond} (Assumption H1). However, the $L$-Lipschitz condition on $\nabla\log p_0$ does not ensure a Lipschitz bound for $\nabla\log p_t$ (see Example~3.4 in \citep{mooney2024global}), and Assumption H1 in \citep{silveribeyond} only guarantees an $\mathcal{O}(1)$ Hessian bound of $\log p_t$ over time, while the Gaussian tail yields an $\mathcal{O}(e^{-t})$ Hessian bound for the modified score function (Corollary~\ref{cor:score lip exp decay bound}) thus leads to improved complexity bounds (see Remark~\ref{rmk:complexity comparsion}). Moreover, the Gaussian tail assumption does not require the target distribution to be log-concave, even far from the origin.

    To demonstrate the \emph{practicality} of the Gaussian tail assumption, we provide two examples. First, in Theorem~\ref{thm:MH}, we show that for any bounded-support target distribution, the density of the forward process at the early stopping time $\delta$ satisfies the assumption with $A=(1-\exp(-\delta))I_d$. Second, the Gaussian tail assumption also holds for certain posterior distributions arising in Bayesian inverse problems; see Theorem~\ref{thm:bayes}.

Under the Gaussian tail Assumption~\ref{ass:Gtail}, we establish bounds on the Hessian and gradient of the transformed potential $\bar{q}(t,x)$ via the heat kernel estimation. 
\begin{theorem}    \label{thm:vHJ hessian bound}
   Under Assumption~\ref{ass:Gtail}, the function  $\bar{q}(t,x)$ in~\eqref{PDE2} satisfies, $\forall t\ge 0$, \begin{align}
         |\nabla \bar{q}(t,\cdot)|_\infty\le |\nabla \bar{q}(0,\cdot)|_\infty, \quad\|\nabla^2 \bar{q}(t,\cdot)\|_\infty \le \|\nabla^2 \bar{q}(0,\cdot)\|_\infty+|\nabla \bar{q}(0,\cdot)|_\infty^2.
    \end{align}
\end{theorem}
Theorem~\ref{thm:vHJ hessian bound} provides uniform in time estimates for the transformed variables in Equation~\eqref{PDE2}. By reversing the change of variables 
\(
(t,x) \rightarrow (t, (\bar{A}_t A^{-1}) e^{t/2} x)
\) 
used to go from equation~\ref{PDE1} to equation~\ref{PDE2}, these uniform bounds translate back to the original coordinates, resulting in the exponential decay estimates for the modified score function stated in the following corollary.

\begin{corollary} \label{cor:score lip exp decay bound}
Define the modified score function as
\begin{align}
    \tilde{s}(t,x) := C \nabla \log p_t(x) + C \A_t^{-1} x.
\end{align}
Suppose that Assumption~\ref{ass:Gtail} holds. Then, $\forall t\ge 0$, the following estimates hold:
\begin{align}
    \|\nabla \tilde{s}(t,\cdot)\|_\infty \le e^{-t} L_0, \quad
    |\tilde{s}(t,\cdot)|_\infty \le e^{-t/2} L_1,
\end{align}
where
\begin{equation*}
\begin{cases}
    K =  \max\{1, \|A C^{-1}\|\},\\
    L_0 = K^2 \big(\|C \nabla^2 h\|_\infty + |\sqrt{C} \nabla h|_\infty^2\big),\\
    L_1 = K \|C\|^{1/2} |\sqrt{C} \nabla h|_\infty.
\end{cases}
\end{equation*}
\end{corollary}
Detailed proofs can be found in Appendices~\ref{proof:vHJ hessian bound} and~\ref{proof:score lip exp decay bound}. To be noted, $K, L_0, L_1$ are bounded dimension-free constants. For later discussion, we also denote,
\begin{equation*}
        \begin{aligned}
        L_2&:=\max \{\|I-CA^{-1}\|,\|AC^{-1}-I\|\}
    \end{aligned}
    \end{equation*}

\subsection{Main Convergence theories}\label{sec:mainbound}
We now turn to the convergence of the scheme~\eqref{eq:explicit solver} to the ground truth target distribution $p_0$ under the Wasserstein-2-distance,
\[ \mathcal{W}_2 (\mu, \nu) = \left(\inf_{\pi} \int | x - y
|^2 d \pi (x, y)\right)^{1/2},  \]
where $\pi$ runs over all measures which has marginals $\mu$ and
$\nu$.
To obtain the convergence results, we assume the following bound on the score approximation at the discretization
points,
\begin{assumption}\label{ass:error} $s_\theta$ is an $\epsilon$ accurate approximation to $s$ on average over the discretization points, i.e.,
    \begin{equation}
     \frac{1}{T} \sum_{k=0}^{N-1} (t_{k+1}-t_k)\e\Big|s(T-t_k,\rx_{T-t_k})-s_\theta(T-t_k,\rx_{T-t_k})\Big|^2  \le \epsilon^2.  
    \end{equation}
\end{assumption}
We would like to point out that, due to the limited access to the score approximation error, when deriving the complexity bounds, we always assume a sufficiently accurate score approximation. 

In addition to the score approximation error (Assumption~\ref{ass:error}), we made the following assumption on the gradient of the learned score to regulate the approximated backward process.
\begin{assumption}\label{ass:appro score lip bound}
    Denote $\tilde{s}_\theta(t,x)=s_\theta(t,x)+C\A_t^{-1}x$, we assume that $\tilde{s}_\theta(t,x)$ shares the same properties as 
$\tilde{s}(t,x)$ in Corollary~\ref{cor:score lip exp decay bound} at the time discretization points, i.e. 
{
\begin{equation}
    \forall k\in\{0,1,\ldots,N-1\},\quad \|\nabla \tilde{s}_\theta(T-t_k.\cdot)\|_\infty \le L_0e^{-T+t_k}.
\end{equation} }
\end{assumption}
The Assumption~\ref{ass:appro score lip bound} can be relaxed to require that the total discrete-time sum of the score gradient is bounded., see Remark~\ref{rmk:relaxed app score}.
Now we present the main theorem in this part.
\begin{theorem}\label{thm:w2 bound}
   Suppose Assumptions~\ref{ass:bound 2 moment}, \ref{ass:Gtail}, \ref{ass:error}, and \ref{ass:appro score lip bound} hold, with the step size $\tau:=\sup_k \{t_{k+1}-t_k\}$ sufficiently small. Then sampling via scheme~\eqref{eq:explicit solver} yields
    \begin{equation}\label{eq:w_2 bound main theorem}
       \mathcal{W}_2^2 (\rp_{0}, \q_{T}) 
    \le K_1\Big(e^{-2T}\left(M_2+\Tr(C)\right)+\epsilon^2T+\Tr(C)\tau^2\Big),
    \end{equation}
    where $K_1=e^{4+3(L_0+L_2)}\max\{(e+1)e,e^3(L_0+L_2)^2\}$ is a dimension-free constant. 
\end{theorem}
Proof see Appendix~\ref{proof:w2 bound}.
As a direct consequence of Theorem~\ref{thm:w2 bound}, 
\begin{corollary}\label{cor:complexity}
  Under the same assumptions as Theorem~\ref{thm:w2 bound}, the following complexity bound guarantees that the distribution $\q_T$ satisfies $\mathcal{W}_2(\rp_0,\q_T) = \mathcal{O}(\epsilon_0)$:
   \begin{align}\label{eq:complexity}
        T&=\mathcal{O}\!\left(\log\frac{M_2+\Tr(C)}{\epsilon_0^2}\right),\quad
        N=\mathcal{O}(\frac{T}{\tau})=\mathcal{O}\!\left(\frac{T}{\epsilon_0}\sqrt{\Tr(C)}\right).
   \end{align}
\end{corollary}
Fixing the second moment $M_2$ and trace of covariance matrix $C$ of base distribution , the complexity bound in Corollary~\ref{cor:complexity} does not depend on the dimension and hence has the potential to extend naturally to infinite-dimensional generative models; see Appendix~\ref{app:infdim}.
 
 Now, we provide two remarks: the first is on the optimality of the result, and the latter is on the relaxation of assumptions on score approximations.
 \begin{remark}\label{rmk:complexity comparsion}
     When assuming the second moment of target distribution $M_2$ and the trace of the diffusion covariance $\Tr(C)$ scale linearly with space dimension $d$, the complexity bound in Corollary~\ref{cor:complexity} is  $\mathcal{O}(\sqrt{d})$ with a logarithmic constant.  In \citep{gao2023wassersteinconvergenceguaranteesgeneral}, Proposition 8 shows that, with the standard Gaussian as the target distribution, such a complexity bound is optimal. Notably, in the line of pursuing complexity bounds under more general assumptions (than log-concaveness), a very recent work \citep{silveribeyond} obtained an $\mathcal{O}(d^2)$ bound\footnote{In arriving at this, we also assume the second moment scales linearly with dimension in their Theorem~3.5.} through the weakly log-concave profile propagation framework established in \citep{saremi2023chain,conforti2024weak,conforti2025projected}. In Table~\ref{tab:complexity}, we provide a non-exhaustive list of complexity bounds for comparison.
 \end{remark}

\begin{remark}\label{rmk:relaxed app score} 
Assumption~\ref{ass:appro score lip bound} can be relaxed as follows.  
We only require that for all $k\in\{0,1,\ldots,N-1\}$,
\begin{align}\label{eq:app score lip sum up a constant}
    \sum_{k=0}^{N-1}(t_{k+1}-t_k)\|\nabla \tilde{s}_\theta(T-t_k,\cdot)\|_\infty \le B,
\end{align}
for some constant $B>0$ independent of $T$.  
Under this condition, Theorem~\ref{thm:w2 bound} still holds with the constant term $K_1$ in \eqref{eq:w_2 bound main theorem} replaced by, $$K_1'=e^{4+3(L_2+eB)}\max\{(e+1)e,e^3(L_0+L_2)^2\}.$$ See Appendix~\ref{app:relaxed app score}.  

\end{remark}

\subsection{Convergence for bounded-support target}\label{sec:manifold}
To illustrate the feasibility of the Gaussian tail Assumption~\ref{ass:Gtail} and the associated complexity bound~\eqref{eq:complexity}, we consider the case when the target distribution has bounded support.
\begin{assumption}\label{ass:boundedsupp}
The target distribution $\rp_0$ is supported on a bounded set, i.e., $\exists R<\infty$, $\diam(\supp \rp_0)\leq R$.
\end{assumption}
The Assumption~\ref{ass:boundedsupp} widely applies to various practical applicative scenario of generative models and can be derived from the manifold hypothesis \citep{bengio2013representation}. To analyze the Lipschitz-ness of the score under the Assumption~\ref{ass:boundedsupp}, we need the following theorem. 
\begin{theorem}\label{thm:MH}
    Let $Q_\sigma=\N(0,\sigma^2I_d)*Q_0$, where $Q_0$ follows Assumption~\ref{ass:boundedsupp}. Then $Q_\sigma$ has smooth density $q_\sigma$ and we define $$g(x):=\log q_\sigma(x) + \frac{|x|^2}{2\sigma^2}.$$ It follows that,
    \begin{align}
        |\nabla g|_\infty\leq \frac{R}{\sigma^2}, \quad
         \|\nabla^2 g\|_\infty\leq\frac{2R^2}{\sigma^4}. \label{eqn:MHineq} 
    \end{align}
\end{theorem}
Proof see Appendix~\ref{proof:MH}.

Similar estimates in Theorem~\ref{thm:MH} can be found in \citep{deconvergence,mooney2024global}. In the current form~\eqref{eqn:MHineq}, we extract the spatial growing part to ensure uniform boundedness in space to be consistent with Assumption~\ref{ass:Gtail}.

 Applying Theorem~\ref{thm:MH} to the forward OU-process~\eqref{eq:forward process} with convariance $C=I_d$ and stopping time $\delta$,
and by Corollary~\ref{cor:score lip exp decay bound}, we get the following Lipschitz bound of the score for bounded-support target distribution.
\begin{corollary}\label{cor:hession bound under manifold ass}
    Suppose that $\rp_0$ follows the Assumption~\ref{ass:boundedsupp}. Let $C=I_d$ and $A=(1-e^{-\delta})I_d$. Then, the corresponding $\A_t=(1-e^{-t})I_d$ and for all $t \ge \delta$, we have, \begin{align*}
        \|\nabla^2\log p_t(x)+\A^{-1}_t\|_\infty\le 3\Big(\frac{R}{1-e^{-\delta}}\Big)^2e^{-t},\quad
        |\nabla \log p_t(x)+\A^{-1}_tx|_\infty \le \frac{R}{1-e^{-\delta}}e^{-\frac{t}{2}}.
    \end{align*}
\end{corollary}
Proof see Appendix~\ref{proof:score lip exp decay bound}. Using the Lipschitz bound in Corollary~\ref{cor:hession bound under manifold ass}, we obtain a Wasserstein-2 distance bound analogous to Theorem~\ref{thm:w2 bound}.
\begin{theorem}
\label{thm:w_2 bound under manifold ass}
     Suppose that $\rp_0$ follows the Assumption~\ref{ass:boundedsupp}, the early stopping time $\delta \le 1$ and the Assumption~\ref{ass:error} holds. Then, sampling via scheme~\eqref{eq:explicit solver} with sufficiently small step size $\tau$ yields,
    \begin{equation} 
      \mathcal{W}^2_2 (\rp_{\delta}, \q_{T-\delta}) 
              \le K_2(R,\delta)\left(e^{-2T+\delta}\left(R^2+d\right)+\epsilon^2T+d\tau^2\right),     
    \end{equation}
    where $K_2(R,\delta)=2\exp \left(7+\frac{12R^2}{(1-e^{-\delta})^2}+\frac{4}{1-e^{-\delta}}\right).$
\end{theorem}
Proof see Appendix~\ref{proof::w_2 bound under manifold ass}.

With the result in Theorem~\ref{thm:w_2 bound under manifold ass}, we can get the complexity bound with early stopping setting under the bounded-support assumption by direct computation.
\begin{corollary}\label{cor:complexity under manifold ass 2}
Under the same assumption as Theorem~\ref{thm:w_2 bound under manifold ass}, achieving a distribution $\q_{T-\delta}$ such that $\mathcal{W}_2(\rp_\delta,\q_{T-\delta}) =\mathcal{O}(\epsilon_0)$ requires:
\begin{align*}
     T=\mathcal{O}\Big(\frac{R^2}{\delta^2}+\log\frac{R^2+d}{\epsilon_0^2}\Big),\quad
     N=\frac{T-\delta}{\tau}=\mathcal{O}\left(\frac{\sqrt{d}T}{\epsilon_0}\exp\left(\frac{2}{1-e^{-\delta}}+\frac{6R^2}{(1-e^{-\delta})^2}\right)\right).
 \end{align*}
 When fixing the early stopping time $\delta$ and support radius $R$, 
 \begin{align}
     N=\mathcal{O}\Big(\frac{\sqrt{d}}{\epsilon_0}\log\frac{R^2+d}{\epsilon_0^2}\Big).
 \end{align}
\end{corollary}
Noticing that,
$\mathcal{W}_2(\rp_\delta,\rp_0)\le \sqrt{\e|\rx_\delta-\rx_0|^2}\le 2 \sqrt{M_0\delta},$
we have the following complexity bound with respect to $\rp_0$.
\begin{corollary}\label{cor:complexity under manifold ass 3}
     Under the same assumption as Theorem~\ref{thm:w_2 bound under manifold ass}, reaching a distribution $\q_{T-\delta}$ such that $\mathcal{W}_2(\rp_0,\q_{T-\delta}) =\mathcal{O}(\epsilon_0)$ requires,
    \begin{align}
        \delta=\mathcal{O}\Big(\frac{\epsilon_0^2}{M_0}\Big),\quad
         \log N= \mathcal{O}\Big(\frac{R^2}{\delta^2}\Big)=\mathcal{O}\Big(\frac{R^2M_0^2}{\epsilon_0^4}\Big).
    \end{align}
\end{corollary}

    The logarithmic complexity $\log N=\mathcal{O}(\frac{R^2}{\delta^2})$ arises jointly from two factors: the $\frac{R^2}{(1-e^{-\delta})^2}$ dependence in the Hessian bound of the early-stopped log density (see Corollary~\ref{cor:hession bound under manifold ass}), and the accumulation of error with respect to the exponential of the Lipschitz-ness of the flow, which is tied to this Hessian bound. To the best of our knowledge, such a dependence on $\delta$ is non-improvable without additional geometric assumptions on the support of $\rp_0$, see \citep{mooney2024global}. Similar exponential dependencies also appear in \citep{de_bortoli_convergence_2023}, while our dependence on $R$ is improved; see Table~\ref{tab:complexity}.

\subsection{Convergence in the Bayesian Inverse problems}\label{sec:beysian}
Another potential application of generative models is to generate the posterior distribution in Bayesian inverse problems. See \citep{stuart2010inverse} for a detailed review. Here, we restrict our theories to the following type of applicative scenario, where we consider a non-linear observation $G \in C^b_2(\R^d,\R^m)$ from the state space $\R^d$ and its observation $y\in \R^m$. We further assume that the prior of the state and the observation noise distribution follow Gaussian-type distributions with   $C$ and $\Sigma$ denoting the covariance matrices, respectively. Then the posterior of the state, also the target distribution of the generative model follows,
\begin{equation}\label{eq:bayePost}
     p_0(x) = D_0\exp\Big(-\frac{|x|_C^2}{2}\Big)\exp\Big(-\frac{|G(x)-y|_\Sigma^2}{2}\Big),
\end{equation} 
where $D_0$ is some normalizing constant. 

 To construct a generative model to sample the posterior, we take the covariance matrix of base distribution in~\eqref{eq:forward process} identical to the prior covariance matrix $C$ in~\eqref{eq:bayePost}. A conditioned score (referred to as the conditional de-noising estimator in the literature \citep{batzolis2021conditional}) is trained with the following loss,
\begin{equation*}
 \e_{p_{t}(x;y)}|s_\theta(t,x;y)-C\nabla_x \log p_{t}(x;y)|^2{,}
\end{equation*}
where $p_{t}$ is then the joint distribution of $(X_t,Y)$ in which $Y$ follows $G(X_0)+\N(0,\Sigma)$. For the generation process of the posterior distribution with observation $y$, we assume the estimated score $s_\theta(t,x;y)$ satisfies Assumption~\ref{ass:error} and~\ref{ass:appro score lip bound}. Then, we have the following theorem.
\begin{theorem}\label{thm:bayes}
    Suppose that the target distribution admits the density in~\eqref{eq:bayePost} and Assumption~\ref{ass:bound 2 moment} holds. Then, sampling via scheme~\eqref{eq:explicit solver} with sufficiently small step size $\tau$ yields,
    \begin{equation}\label{est:bayes} 
           \mathcal{W}_2^2 \big(\rp_0(\cdot;y), \q_{T}(\cdot;y)\big) 
              \le K_3\Big(e^{-2T}\big(M_2+\Tr(C)\big)+\epsilon^2T+\Tr(C)\tau^2\Big),
    \end{equation}
where $K_3$ is a dimensionless constant determined by $(\|C\|,\|\Sigma\|, G,y)$, the explicit expression is provided in Table~\ref{tab:constants}.
\end{theorem} 
Proof see Appendix~\ref{proof:bayes}.

\bibliography{ref}
\bibliographystyle{plain}

\newpage
\appendix \onecolumn
\section*{Appendix}
 The appendix consists of three parts.  In Section~\ref{app:proofs}, we present all the detailed proofs. In Section~\ref{app:infdim}, we discuss generalizing our theories to infinite dimensions. In Section~\ref{app:constants}, we provide the explicit expressions of the constants appearing in the context.

\section{Proofs of Theorems}\label{app:proofs}
Here we present detailed proofs. 
\subsection{Proof of Theorem~\ref{thm:vHJ hessian bound} (heat kernel estimation)}\label{proof:vHJ hessian bound}
\begin{proof}
    Consider the solution of~\eqref{eq:PDE3} given by~\eqref{eq:solution of PDE3},
    \begin{align*}
       \bar{p}(t,x)= \frac{1}{(2\pi)^n}\int _{\mathbb{R}^n}\frac{1}{\sqrt{\det B(t)}}\exp\Big({\frac{-|x-y|_{B(t)}^2}{2}}\Big)\exp\big(-\bar{h}(y)\big)dy ,\quad (t,x)\in(0, \infty)\times \mathbb{R}^n.
    \end{align*}
So $\bar{q}(t,x)=-\log \bar{p}(t,x)$ satisfies,
\begin{align*}
    \nabla_x \bar{q}(t,x) &=-\frac{\nabla_x \bar{p}(t,x)}{\bar{p}(t,x)}=-\frac{\int _{\mathbb{R}^n}\big(\nabla _x\exp({\frac{-|x-y|_{B(t)}^2}{2}})\big)\exp\big(-\bar{h}(y)\big)dy}{\int _{\mathbb{R}^n}\exp({\frac{-|x-y|_{B(t)}^2}{2}})\exp\big(-\bar{h}(y)\big)dy}\\
    &=\frac{\int _{\mathbb{R}^n}\big(\nabla _y\exp({\frac{-|x-y|_{B(t)}^2}{2}})\big)\exp\big(-\bar{h}(y)\big)dy}{\int _{\mathbb{R}^n}\exp({\frac{-|x-y|_{B(t)}^2}{2}})\exp\big(-\bar{h}(y)\big)dy} \\
    &=\frac{\int _{\mathbb{R}^n}\big(\nabla _y \bar{h}(y)\big)\exp({\frac{-|x-y|_{B(t)}^2}{2}})\exp\big(-\bar{h}(y)\big)dy}{\int _{\mathbb{R}^n}\exp({\frac{-|x-y|_{B(t)}^2}{2}})\exp\big(-\bar{h}(y)\big)dy}.
\end{align*}
Here, the third line is derived from the integration by parts formula.

Since $\exp({\frac{-|x-y|_{B(t)}^2}{2}})\exp\big(-\bar{h}(y)\big) \ge 0$, taking absolute value we get $|\nabla_x \bar{q}(t,x)| \le |\nabla \bar{h}|_\infty$, thus, \begin{align*}
    |\nabla \bar{q}(t,\cdot)|_\infty \le |\nabla \bar{h}|_\infty=|\nabla \bar{q}(0,\cdot)|_\infty.
\end{align*} 
{For any unit direction $z$, we denote $\nabla_z$ as taking derivative along that direction and $\nabla_z$ as taking derivative twice along that direction, i.e., for a given $C^2$ function $f$,
\begin{align*}
   \nabla_z f(x)=\langle z, \nabla f(x)\rangle, \quad \nabla_z^2f(x)= \langle z, \nabla^2f(x) z\rangle,
\end{align*}
where $\nabla f(x)$ is the gradient of $f$ and $\nabla^2 f(x)$ is the Hessian matrix of $f$ at $x$.
Using the same method as above, we get,\begin{align*}
    \nabla^2_z  \bar{q}(t,x)&=\frac{\int _{\mathbb{R}^n}\big(\nabla _z^2 \bar{h}-(\nabla_z\bar{h})^2\big)(y)\exp({\frac{-|x-y|_{B(t)}^2}{2}})\exp\big(-\bar{h}(y)\big)dy}{\int _{\mathbb{R}^n}\exp({\frac{-|x-y|_{B(t)}^2}{2}})\exp\big(-\bar{h}(y)\big)dy}\\&+\Big(\frac{\int _{\mathbb{R}^n}\big(\nabla _z \bar{h}(y)\big)\exp({\frac{-|x-y|_{B(t)}^2}{2}})\exp\big(-\bar{h}(y)\big)dy}{\int _{\mathbb{R}^n}\exp({\frac{-|x-y|_{B(t)}^2}{2}})\exp\big(-\bar{h}(y)\big)dy}\Big)^2.
\end{align*}
Taking absolute value again, we get, \begin{align*}
    | \nabla^2_z \bar{q}(t,x)| \le |\nabla _z^2 \bar{h}|_\infty+|\nabla_z\bar{h}|_\infty^2\le \|\nabla^2\bar{h}\|_\infty+|\nabla \bar{h}|^2_\infty,\quad \forall x \in \mathbb{R}^n,
\end{align*}
thus,\begin{align*}
     \|\nabla^2\bar{q}(t,x)\|=\sup_{|z|=1}|\nabla_z^2\bar{q}(t,x)|\le|\nabla^2\bar{h}\|_\infty+|\nabla \bar{h}|^2_\infty, \quad \forall x \in R^n.
\end{align*}
Since the bound for $ \|\nabla^2\bar{q}(t,x)\|$ does not rely on $x$, we get, \begin{align}
    \|\nabla^2\bar{q}(t,\cdot)\|_\infty \le \|\nabla^2\bar{h}\|_\infty+|\nabla \bar{h}|_\infty^2=\|\nabla^2 \bar{q}(0,\cdot)\|_\infty+|\nabla \bar{q}(0,\cdot)|_\infty^2.
\end{align}}
\end{proof}
The above analysis is developed to facilitate the case $C\neq I_d$; other standard PDE approaches, like the classical Bernstein method, can also be applied for the isotropic case, i.e., $C=I_d$.
\subsection{Proof of Corollary~\ref{cor:score lip exp decay bound} and Corollary~\ref{cor:hession bound under manifold ass}}\label{proof:score lip exp decay bound}
\begin{proof}
  Recall that $\bar{q}(t,x)=q(t,K(t)^{-1}\sqrt{C}x)+f(t)$, so, 
   \begin{align*}
       K(t)\sqrt{C^{-1}}\nabla \bar{q}(t,x)=& \nabla q (t,K(t)^{-1}\sqrt{C}x),\\
       K(t)^2C^{-1}\nabla^2 \bar{q}(t,x)=&\nabla^2 q (t,K(t)^{-1}\sqrt{C}x).
   \end{align*}
Notice that~$$
        \tilde{s}(t,x)=C\nabla q(t,x), \quad 
        \nabla\tilde{s}(t,x)= C\nabla^2 q(t,x).$$
Hence,
    \begin{align*}
        |\tilde{s}(t,\cdot)|_\infty=&|C\nabla q(t,\cdot)|_\infty=|K(t)\sqrt{C}\nabla \bar{q}(t,\cdot)|_\infty \le e^{-\frac{t}{2}}\|e^{\frac{t}{2}}K(t)\|\|\sqrt{C}\||\nabla \bar{q}(t,\cdot)|_\infty,\\ 
        \|\nabla \tilde{s}(t,\cdot)\|_\infty =&\|C\nabla^2q(t,\cdot)\|_\infty=\|K(t)^2\nabla^2 \bar{q}(t,\cdot)\|_\infty \le e^{-t}\|e^{\frac{t}{2}}K(t)\|^2\|\nabla^2 \bar{q}(t,\cdot)\|_\infty.
    \end{align*}
Define the constant, $$K:=\sup_{t \ge 0}\|e^{\frac{t}{2}}K(t)\|=\sup_{t\ge 0}\|A\A^{-1}_t\|=\max\{1, \|A C^{-1}\|\}.$$ By Theorem~\ref{thm:vHJ hessian bound} and the initial value $\bar{q}(0,x)=h(\sqrt{C}x)$, we obtain,
\begin{equation}
    \begin{aligned}
    |\tilde{s}(t,\cdot)|\le& e^{-\frac{t}{2}}K \|C\|^{\frac{1}{2}}|\sqrt{C}\nabla h|_\infty,\\
        \|\nabla \tilde{s}(t,\cdot)\|_\infty \le& e^{-t}K^2\Big(\|C\nabla h\|_\infty+|\sqrt{C}\nabla h|_\infty^2\Big).
    \end{aligned}
\end{equation} 
Let, \begin{equation*}
\begin{aligned}
    L_0:=&K^2\Big(\|C\nabla h\|_\infty+|\sqrt{C}\nabla h|_\infty^2\Big),\\
    L_1:=&K \|C\|^{\frac{1}{2}}|\sqrt{C}\nabla h|_\infty,
\end{aligned}
\end{equation*}then we get the results in Corollary~\ref{cor:score lip exp decay bound}.

Now we turn to proof of Corollary~\ref{cor:hession bound under manifold ass}. Recall that, in Corollary~\ref{cor:hession bound under manifold ass},  we have,\begin{equation}\nonumber
\begin{cases}
    C=I_d,\\
        A=(1-e^{-\delta})I_d ,\\
        \A_t=(1-e^{-t})I_d.
\end{cases}
\end{equation}
And the distribution of the forward OU process at time~$\delta$ is given by  \begin{align*}
        \rp_\delta=\N\big(0,(1-e^{-\delta})I_d\big)*P_{0,\delta},
\end{align*}
where $P_{0,\delta}:=Law(e^{-\frac{\delta}{2}}X_0),\quad X_0\sim \rp_0$ and $P_{0,\delta}$ satisfies $\diam (\supp P_{0,\delta})\le e^{-\frac{\delta}{2}}R$.
Then, by Theorem~\ref{thm:MH}, the corresponding $h(x)=\log p_\delta(x)+\frac{|x|^2}{2(1-e^{\delta})}$ satisfies,
\begin{equation*}
    |\nabla h|_\infty \le \frac{Re^{-\frac{\delta}{2}}}{1-e^{-\delta}}, \quad \|\nabla^2 h\| \le 2\frac{R^2e^{-\delta}}{(1-e^{-\delta})^2}.
\end{equation*}
The constants then can be computed directly,
\begin{equation*}
    \begin{cases}
        K=\sup_{t \ge \delta}\|A\A^{-1}_t\|=1,\\
        L_0= K^2(\|\nabla h\|_\infty+|\nabla h|_\infty^2)\le 3\frac{R^2e^{-\delta}}{(1-e^{-\delta})^2},\\
        L_1=K|\nabla h|_\infty \le \frac{Re^{-\frac{\delta}{2}}}{1-e^{-\delta}}.
    \end{cases}
\end{equation*}
Then, by Corollary~\ref{cor:score lip exp decay bound}, for all~$ t \ge \delta$,
\begin{equation}
    \begin{aligned}
    \|\nabla^2 \log p_t(x)+\A^{-1}_t\|_\infty \le& L_0e^{-(t-\delta)}=3\frac{R^2}{(1-e^{-\delta})^2}e^{-t},\\
    |\nabla^2 \log p_t(x)+\A^{-1}_tx|_\infty \le & L_1e^{-\frac{t-\delta}{2}}=\frac{R}{1-e^{-\delta}}e^{-\frac{t}{2}}.
\end{aligned}
\end{equation}
\end{proof}

\subsection{Proof of Theorem~\ref{thm:w2 bound}}\label{proof:w2 bound}

We first derive some estimates that will be useful in the proof.
Since \begin{equation*}
        \begin{aligned}
        \sup_{t\ge 0}\|e^t(I-C\A^{-1}_t)\|=&\sup_{t\ge 0}\|(A-C)\Big(Ae^{-t}+C(1-e^{-t})\Big)^{-1}\|\\
        =& \max \{\|I-CA^{-1}\|,\|AC^{-1}-I\|\},\\
        =&L_2.
    \end{aligned}
    \end{equation*}
 we have the following estimate, 
 \begin{equation*}\label{computation:L_2,L_3}
        \|I-C\A^{-1}_t\|
        \le L_2e^{-t}, \quad \forall t\ge0.
    \end{equation*}

Recall from Corollary~\ref{cor:score lip exp decay bound} that, for all~$t \ge 0$,
    \begin{equation*}
        \|\nabla \tilde{s}(t,\cdot)\|_\infty \le L_0e^{-t},\quad 
        |\tilde{s}(t,\cdot)|_\infty \le L_1e^{-\frac{t}{2}},
    \end{equation*}
then,
\begin{align}\label{computation:lip of s+I}
    \|\nabla s(t,\cdot)+I\|_\infty\le \|\nabla \tilde{s}(t,\cdot)\|_\infty+\|I-C\A_t^{-1}\|\le (L_0+L_2)e^{-t}.
\end{align}
Similarly, under Assumption~\ref{ass:appro score lip bound}, we have, 
\begin{align}\label{computation:lip of s_theta+I}
   \forall k\in \{0,1,...,N-1\},\quad \|\nabla \hat{s}_\theta(T-t_k,\cdot)\|_\infty=\|\nabla s_\theta(T-t_k,\cdot)+I\|_\infty\le (L_0+L_2)e^{-T+t_k}.
\end{align}
Throughout this proof, we set the step size
\[
\tau := \sup_{k}(t_{k+1}-t_k) \le \min\Big\{1,\frac{1}{(L_0+L_2)e}\Big\},
\]
and can validate that
\begin{align}\label{eq:small tau}
    \forall r>0,\quad\frac{e^{r\tau}-1}{r}\le e^r\tau ,\quad  
\textrm{and} \quad (L_0+L_2)e\tau \le 1.
\end{align}

Here, we present a lemma related to convergence to equilibrium for the OU process.
\begin{lemma}[Theorem 23.26 \citep{villani2009optimal}]
    \label{lem:wgf} Let $V$ be $\lambda$-uniformly convex $C^2$ potential. Consider the Langevin process,
    \begin{align*}
        dX_t =-\frac{1}{2}\nabla V(X_t)dt + \sqrt{C}dW_t,
    \end{align*}
    with two initial measure $\mu_0$ and $\nu_0$. 
    \begin{align*}
         \mathcal{W}_2  (\mu_t,\nu_t)\leq  \mathcal{W}_2  (\mu_0,\nu_0)e^{- \frac{\lambda t}{2}}.
    \end{align*}
\end{lemma}
The convergence of OU is a direct consequence, with $\lambda=1$.
We also need the following martingale property for the score function.
\begin{lemma}\label{lem:rever martingale}
    The quantity $e^{-\frac{t}{2}}\left(s(T-t,\lx_t)+\lx_t\right)$ is a martingale, moreover,
    \begin{align}
        de^{-\frac{t}{2}}\left(s(T-t,\lx_t)+\lx_t\right)=e^{-\frac{t}{2}}\left(I+\nabla s(T-t,\lx_t)\right)\sqrt{C}d\tilde{B}_t.
    \end{align}
\end{lemma}
\begin{proof}
By It\^{o}'s formula and \eqref{eq:backward process},
\begin{align*}
    &d s(T-t,\lx_t)=-\partial_ts(T-t,\lx_t)dt+\nabla s(T-t,\lx_t)d\lx_t+\frac{1}{2}C\nabla \left(\nabla\cdot s(T-t,\lx_t)\right)dt\\
    =&-\partial_ts(T-t,\lx_t)+\frac{1}{2}\left(\nabla s(T-t,\lx_t)\right )\lx_tdt+\left(\nabla s(T-t,\lx_t)\right ) s(T-t,\lx_t)dt\\
    &+\left(\nabla s(T-t,\lx_t)\right )\sqrt{C}dB_t+\frac{1}{2}C\nabla \left(\nabla\cdot s(T-t,\lx_t)\right)dt.
\end{align*}
From the Fokker-Planck equation,
$   \partial_tp=\frac{1}{2}\left(\nabla\cdot (xp+C\nabla p)\right)$, 
we have,
\begin{align*}
    \partial_t s=&\partial_t C\nabla \log p=C\left(\frac{\nabla \partial_tp}{p}-\frac{\partial_t p}{p^2}\nabla p\right)\\
    =&\frac{1}{2}C\left( \frac{\nabla \left(\nabla\cdot\left(xp+C\nabla p\right)\right)}{p}-\frac{\nabla\cdot\left(xp+C\nabla p\right)}{p^2}\nabla p\right)\\
    =&\frac{1}{2}C\left(\frac{\nabla\left(dp+x\cdot\nabla p+\nabla \cdot p\left(C\nabla\log p\right)\right)}{p}-\frac{dp+x\cdot\nabla p+\nabla \cdot \left(pC\nabla \log p\right)}{p^2}\nabla p\right)\\
    =&\frac{1}{2}C\left(d\nabla \log p+\frac{\nabla \left(x\cdot p\nabla \log p+\left(\nabla p\right) \cdot C\nabla \log p+p\nabla \cdot C \nabla \log p\right)}{p}\right.\\
    &\left.-d\nabla \log p-\left(x\cdot \nabla\log p\right)\nabla \log p-\frac{\left(\nabla p\right)\cdot C\nabla\log p+p\nabla \cdot C\nabla \log p }{p}\nabla \log p \right)\\
    =&\frac{1}{2}C\left(\nabla \log p+ \left(x\cdot \nabla \log p\right)\nabla \log p+\left(\nabla ^2 \log p\right)x+\frac{\nabla \left(\left(p\nabla \log p\right)\cdot C\nabla
     \log p\right)}{p}\right.\\
     &\left.+\left(\nabla \cdot C\nabla \log p \right)\nabla \log p+\nabla \left(\nabla \cdot C\nabla \log p\right)-\left(x\cdot \nabla \log p\right)\nabla \log p \right.\\
     &\left.-\left(\nabla \log p \cdot C\nabla \log p \right)\nabla \log p-\left(\nabla \cdot C\nabla \log p \right)\nabla \log p\right.\bigg)\\
     =&\frac{1}{2}C\left( \nabla \log p+\left(\nabla^2 \log p\right)x+\left(\nabla \log p \cdot C\nabla \log p\right)\nabla \log p+2\left(\nabla ^2\log p \right)C\nabla \log p\right.\\
     &\left.+\nabla \left(\nabla \cdot C\nabla \log p\right)-\left(\nabla \log p \cdot C\nabla \log p\right)\nabla \log p\right)\\
     =&\frac{1}{2}C\left( \nabla \log p+\left(\nabla^2 \log p\right)x+2\left(\nabla ^2\log p \right)C\nabla \log p+\nabla \left(\nabla \cdot C\nabla \log p\right)\right)\\
     =&\frac{1}{2}\left(s+\left (\nabla s\right )x+2\left(\nabla s\right)s+C\nabla \left (\nabla \cdot s\right )\right).
\end{align*}
Thus,
\begin{align*}
    d s(T-t,\lx_t)=&-\frac{1}{2}s(T-t,\lx_t)dt+\left(\nabla s(T-t,\lx_t)\right )\sqrt{C}dB_t.
\end{align*}
Combining with It\^{o}'s formula applied to $e^{-\frac{t}{2}}\left(s(T-t,\lx_t)+\lx_t\right)$, we obtain,
\begin{align*}
     &de^{-\frac{t}{2}}\left(s(T-t,\lx_t)+\lx_t\right)\\=&de^{-\frac{t}{2}}\left(s(T-t,\lx_t)\right)+de^{-\frac{t}{2}}\lx_t\\
     =&e^{-\frac{t}{2}}ds(T-t,\lx_t)-\frac{1}{2}e^{-\frac{t}{2}} s(T-t,\lx_t)dt-\frac{1}{2}e^{-\frac{t}{2}}\lx_tdt+e^{-\frac{t}{2}}d\lx_t\\
     =&e^{-\frac{t}{2}}\nabla s(T-t,\lx_t)\sqrt{C}d\tilde{B}_t-e^{-\frac{t}{2}}(\frac{1}{2}\lx_t+s(T-t,\lx_t))dt+e^{-\frac{t}{2}}d\lx_t\\
     =&e^{-\frac{t}{2}}\left(I+\nabla s(T-t,\lx_t)\right)\sqrt{C}d\tilde{B}_t.
\end{align*}
\end{proof}

\paragraph{Proof of Theorem~\ref{thm:w2 bound}}
For simplicity, we denote, $$\hat{s}(t,x):=s(t,x)+x,\quad \hat{s}_\theta(t,x):=s_\theta(t,x)+x.$$
By It\^{o}'s formula and \eqref{eq:backward process}, we have, $\forall k \in \{0,1,...,N-1\}\ \mathrm{and}\ \forall t \in [t_k, t_{k+1})$,
\begin{align*}
    d e^{\frac{t}{2}}\lx_t=&\frac{1}{2}e^{\frac{t}{2}}\lx_tdt+e^{\frac{t}{2}}d\lx_t=e^{\frac{t}{2}}\hat{s}(T-t,\lx_t)dt+e^{\frac{t}{2}}\sqrt{C}d\tilde{B}_t,
\end{align*}
thus,
\begin{align*}
    e^{\frac{t_{k+1}}{2}}\lx_{t_{k+1}}=e^{\frac{t_{k}}{2}}\lx_{t_{k}}+\int _{t_k}^{t_{k+1}}e^{\frac{t}{2}}\hat{s}(T-t,\lx_t)dt+\sqrt{e^{t_{k+1}-t_k}}\tilde{z},
\end{align*}
where $\tilde{z}_k\sim \N(0,C)$.
Now, we couple two processed $\lx_t$ and  $\ly_t $ with the same Brownian motion, i.e., let $\bar{z}_k=\tilde{z}_k$ in the sampling scheme~\eqref{eq:explicit solver}, then, 
\begin{align*}
    e^{\frac{t_{k+1}}{2}}(\lx_{t_{k+1}}-\ly_{t_{k+1}})=&e^{\frac{t_{k}}{2}}(\lx_{t_{k}}-\ly_{t_{k}})+\int _{t_k}^{t_{k+1}}\left(e^{\frac{t}{2}}\hat{s}(T-t,\lx_t)-e^{t-\frac{t_k}{2}}\hat{s}_\theta(T-t_k,\ly_{t_k})\right)dt\\
    =&e^{\frac{t_{k}}{2}}(\lx_{t_{k}}-\ly_{t_{k}})+\int _{t_k}^{t_{k+1}}e^{t}\left(e^{-\frac{t}{2}}\hat{s}(T-t,\lx_t)-e^{-\frac{t_k}{2}}\hat{s}(T-t_k,\lx_{t_k})\right)dt\\
    &+\int _{t_k}^{t_{k+1}}e^{t-\frac{t_k}{2}}\left(\hat{s}(T-t_k,\lx_{t_k})-\hat{s}_\theta(T-t_k,\lx_{t_k})\right)dt\\
   & +\int _{t_k}^{t_{k+1}}e^{t-\frac{t_k}{2}}\left(\hat{s}_\theta(T-t_k,\lx_{t_k})-\hat{s}_\theta(T-t_k,\ly_{t_k})\right)dt,
\end{align*}
Recall in Lemma~\ref{lem:rever martingale},
\begin{align*}
    e^{-\frac{t}{2}}\hat{s}(T-t,\lx_t)-e^{-\frac{t_k}{2}}\hat{s}(T-t_k,\lx_{t_k})=\int_{t_k}^te^{-\frac{s}{2}}\nabla \hat{s}(T-s,\lx_t)\sqrt{C}d\tilde{B}_s,
\end{align*}
thus,
\begin{align*}
    \e \left| e^{\frac{t_{k+1}}{2}}(\lx_{t_{k+1}}-\ly_{t_{k+1}})\right|^2=&\e\Bigg| e^{\frac{t_{k}}{2}}(\lx_{t_{k}}-\ly_{t_{k}})\\
    &\quad+\int _{t_k}^{t_{k+1}}e^{t-\frac{t_k}{2}}\left(\hat{s}_\theta(T-t_k,\lx_{t_k})-\hat{s}_\theta(T-t_k,\ly_{t_k})\right)dt\\
    &\quad+\int _{t_k}^{t_{k+1}}e^{t-\frac{t_k}{2}}\left(\hat{s}(T-t_{k},\lx_{t_k})-\hat{s}_\theta(T-t_k,\lx_{t_k})\right)dt\Bigg|^2\\
    &+\e \Bigg |\int_{t_k}^{t_{k+1}}e^t\int_{t_k}^te^{-\frac{s}{2}}\nabla \hat{s}(T-s,\lx_t)\sqrt{C}d\tilde{B}_sdt\Bigg|^2\\
    :=&I_1^k+I_2^k.
\end{align*}
By Cauchy's inequality,
\begin{align*}
    I_2^k\le& \e\left|\int_{t_k}^{t_{k+1}}e^{2t}dt\int_{t_k}^{t_{k+1}}|\int_{t_k}^te^{-\frac{s}{2}}\nabla \hat{s}(T-s,\lx_t)\sqrt{C}d\tilde{B}_s|^2dt\right|\\
    =&\frac{e^{2t_{k+1}}-e^{{2t_k}}}{2}\int_{t_k}^{t_{k+1}}\e|\int_{t_k}^te^{-\frac{s}{2}}\nabla \hat{s}(T-s,\lx_t)\sqrt{C}d\tilde{B}_s|^2 dt.
\end{align*}
By It\^{o}'s isometry and~\eqref{computation:lip of s+I},
\begin{align*}
    \e|\int_{t_k}^te^{-\frac{s}{2}}\nabla \hat{s}(T-s,\lx_t)\sqrt{C}d\tilde{B}_s|^2&\le \int_{t_k}^te^{-s}(L_0+L_2)^2e^{-2T+2s}\Tr(C)ds\\
    &\le(L_0+L_2)^2\Tr(C)e^{-2T}(e^{t_{k+1}}-e^{t_k}).
\end{align*}
With sufficiently small $\tau$ satisfying~\eqref{eq:small tau}, we have,
\begin{align}\label{computation:I_2^k}
    I_2^k\le&e^3(L_0+L_2)^2\Tr(C)e^{-2T+3t_k}\tau^2(t_{k+1}-t_k).
\end{align}
By mean inequality, for any positive number $f_k$, we have
\begin{align}
    I_1^k\le& (1+f_k)\e \Bigg| e^{\frac{t_{k}}{2}}(\lx_{t_{k}}-\ly_{t_{k}})+\int _{t_k}^{t_{k+1}}e^{t-\frac{t_k}{2}}\left(\hat{s}_\theta(T-t_k,\lx_{t_k})-\hat{s}_\theta(T-t_k,\ly_{t_k})\right)dt\Bigg|^2\nonumber\\
    &+(1+f_k^{-1})\e \Bigg |\int _{t_k}^{t_{k+1}}e^{t-\frac{t_k}{2}}\left(\hat{s}(T-t_{k},\lx_{t_k})-\hat{s}_\theta(T-t_k,\lx_{t_k})\right)dt\Bigg |^2\label{I1k decomposition}
\end{align}
By~\eqref{computation:lip of s_theta+I} and the condition~\eqref{eq:small tau} on $\tau$, the first term of \eqref{I1k decomposition} satisfies,
\begin{align*}
    &\e \Bigg| e^{\frac{t_{k}}{2}}(\lx_{t_{k}}-\ly_{t_{k}})+\int _{t_k}^{t_{k+1}}e^{t-\frac{t_k}{2}}\left(\hat{s}_\theta(T-t_k,\lx_{t_k})-\hat{s}_\theta(T-t_k,\ly_{t_k})\right)dt\Bigg|^2\\
    \le&\e\Bigg| e^{\frac{t_{k}}{2}}(\lx_{t_{k}}-\ly_{t_{k}})+\int _{t_k}^{t_{k+1}}e^{t-\frac{t_k}{2}}(L_0+L_2)e^{-T+t_k}(\lx_{t_k}-\ly_{t_k})dt\Bigg|^2\\   
    \le&\left(1+(L_0+L_2)e^{-T}\int_{t_k}^{t_{k+1}}e^tdt\right)^2\e| e^{\frac{t_{k}}{2}}(\lx_{t_{k}}-\ly_{t_{k}})|^2\\
    \le & \left(1+3(L_0+L_2)e^{-T}(e^{t_{k+1}}-e^{t_k})\right)\e|e^{\frac{t_{k}}{2}}(\lx_{t_{k}}-\ly_{t_{k}})|^2,
\end{align*}
Denoting $$\epsilon_k:=\sqrt{\e|s(T-t_k,\lx_{t_k})-s_\theta(T-t_k,\lx_{t_k})|^2},$$ the second term of \eqref{I1k decomposition} satisfies,
\begin{align*}
    &\e \Bigg |\int _{t_k}^{t_{k+1}}e^{t-\frac{t_k}{2}}\left(\hat{s}(T-t_{k},\lx_{t_k})-\hat{s}_\theta(T-t_k,\lx_{t_k})\right)dt\Bigg |^2\\
    =&(e^{t_{k+1}}-e^{t_k})^2e^{-t_k}\e |s(T-t_{k},\lx_{t_k})-s_\theta(T-t_k,\lx_{t_k})|^2\\
    =& (e^{t_{k+1}}-e^{t_k})^2e^{-t_k}\epsilon_k^2.
\end{align*}
Now, for all $f_k>0$, we have,
\begin{align*}
    I_1^k\le& (1+f_k)\left(1+3(L_0+L_2)e^{-T}(e^{t_{k+1}}-e^{t_k})\right)\e|e^{\frac{t_{k}}{2}}(\lx_{t_{k}}-\ly_{t_{k}})|^2\\
    &+(1+f_k^{-1})(e^{t_{k+1}}-e^{t_k})^2e^{-t_k}\epsilon_k^2.
\end{align*}
Let,  \begin{align*}
    f_k=e^{-T}(e^{t_{k+1}}-e^{t_k})\le e^\tau -1\le e,
\end{align*}
then, by the condition~\eqref{eq:small tau} on $\tau$,
\begin{align*}
    (1+f_k)(1+3(L_0+L_2)e^{-T}(e^{t_{k+1}}-e^{t_k}))
    \le&1+f_k+3(L_0+L_2)e^{-T}(e^{t_{k+1}}-e^{t_k})+3f_k\\
    =&1+\Big(4+3(L_0+L_2)\Big)e^{-T}(e^{t_{k+1}}-e^{t_k})\\
    (1+f_k^{-1})(e^{t_{k+1}}-e^{t_k})^2e^{-t_k}
    \le& (e+1)f_k^{-1}(e^{t_{k+1}}-e^{t_k})^2e^{-t_k}\\
    =& (e+1)e^Te(t_{k+1}-t_k),
\end{align*}
thus,
\begin{equation}\label{computation:I_1^k}
    \begin{aligned}
    I_1^k\le& \Big(1+\big(4+3(L_0+L_2)\big)e^{-T}(e^{t_{k+1}}-e^{t_k})\Big)\e|e^{\frac{t_{k}}{2}}(\lx_{t_{k}}-\ly_{t_{k}})|^2\\
    &+(e+1)e^{T}\epsilon_k^2e(t_{k+1}-t_k).
\end{aligned}
\end{equation}

Denoting $E(k)=\e|e^{t_k}(\lx_{t_k}-\ly_{t_k})|^2$ and combining the estimates~\eqref{computation:I_2^k},~\eqref{computation:I_1^k} for $I_2^k$ and $I_1^k$, we obtain,
\begin{align*}
    E(k+1)\le& \Big(1+\big(4+3(L_0+L_2)\big)e^{-T}(e^{t_{k+1}}-e^{t_k})\Big)E(k)+(e+1)e^{T}\epsilon_k^2e(t_{k+1}-t_k)\\&+e^3(L_0+L_2)^2\Tr(C)e^{-2T+3t_k}\tau^2(t_{k+1}-t_k)\\
    \le&\exp \Big(\big(4+3(L_0+L_2)\big)e^{-T+t_{k+1}}\Big)\Bigg(E(k)\exp\Big(-\big(4+3(L_0+L_2)\big)e^{-T+t_{k}}\Big)\\
    &+ (e+1)e^{T}\epsilon_k^2e(t_{k+1}-t_k)+ e^3(L_0+L_2)^2\Tr(C)e^{-2T}\tau^2(e^{3t_{k+1}}-e^{3t_k})\Bigg),
\end{align*}
where we used $\exp \Big(\big(4+3(L_0+L_2)\big)e^{-T+t_{k+1}}\Big)\geq1$. Equivalently, we have
\begin{equation*}
    \begin{aligned}
    &E(k+1)\exp\Big(-\big(4+3(L_0+L_2)\big)e^{-T+t_{k+1}}\Big)-E(k)\exp\Big(-\big(4+3(L_0+L_2)\big)e^{-T+t_{k}}\Big)\\
    \le& (e+1)e^{T}\epsilon_k^2e(t_{k+1}-t_k)+ e^3(L_0+L_2)^2\Tr(C)e^{-2T}\tau^2(e^{3t_{k+1}}-e^{3t_k}).
\end{aligned}
\end{equation*}
Summing up over $k$ and using the condition in Assumption~\ref{ass:error}, we obtain,
\begin{align*}
    &\e|e^{\frac{T}{2}}(\lx_T-\ly_T)|^2\exp\Big(-4-3(L_0+L_2)\Big)- \e|(\lx_0-\ly_0)|^2\exp\Big(-\big(4+3(L_0+L_2)\big)e^{-T}\Big)\\
    \le&(e+1)e^{T+1}\sum_{k=0}^{N-1}\epsilon_k^2(t_{k+1}-t_k)+ e^3(L_0+L_2)^2\Tr(C)e^{-2T}\tau^2(e^{3T}-1)\\
    \le& (e+1)e^{T+1}\epsilon^2T+ e^3(L_0+L_2)^2\Tr(C)e^{-2T}\tau^2(e^{3T}-1),
\end{align*}
Denoting $K_1:=e^{4+3(L_0+L_2)}\max\{(e+1)e,e^3(L_0+L_2)^2\}$, we have the following bound for $\e|\lx_T-\ly_T|^2$,
\begin{align*}
    \e|\lx_T-\ly_T|^2\le K_1\left(e^{-T}\e|\lx_0-\ly_0|^2+\epsilon^2T+\Tr(C)\tau^2\right).
\end{align*}
Now we pick a $\xi$-optimal coupling of $\lx_{0} \sim \rp_T$ and $\ly_{0} \sim \q_0=\gamma_C$  in the  Wasserstein distance, i.e. \begin{align*}
    \e|\lx_{0}-\ly_{0}|^2 \le \mathcal{W}_2^2 (\rp_{T}, \q_0)+\xi,
\end{align*}
and obtain,
\begin{align*}
    \mathcal{W}^2_2 (\rp_{0}, \q_T) \le \e|\lx_T-\ly_T|^2 \le K_1(e^{-T}(\mathcal{W}^2_2 (\rp_{T}, \q_0)+\xi)+\epsilon^2T+\Tr(C)\tau^2).
\end{align*}
Since $\xi$ is arbitrary, the bound will be,
\begin{align*}
    \mathcal{W}_2 (\rp_{0}, \q_T) \le K_1(e^{-T}\mathcal{W}^2_2 (\rp_{T}, \q_0)+\epsilon^2T+\Tr(C)\tau^2).
\end{align*}
Noticing that,
\begin{align*}
   \mathcal{W}_2^2 (\rp_{0}, \gamma_C) \le \e_{\rp_{0}\otimes\gamma_C} |x-y|^2=\e_{\rp_{0}}|x|^2+\e_{\gamma_C}|y|^2=M_2+\Tr(C),
\end{align*}
and by Lemma \ref{lem:wgf}, 
\begin{align*}
     \mathcal{W}_2^2 (\rp_{T}, \q_T)=\mathcal{W}_2^2 (\rp_{T}, \gamma_C) \le e^{-T}\mathcal{W}_2^2 (\rp_{0}, \gamma_C).
\end{align*}
We get,
\begin{equation}\label{eq:consequence of w_2 bound proof}
      \mathcal{W}_2^2 (\rp_{0}, \q_{T}) 
    \le K_1\left(e^{-2T}\left(M_2+\Tr(C)\right)+\epsilon^2T+\Tr(C)\tau^2\right).
\end{equation}

\begin{remark}\label{rmk:w_2 under early stop setting}
    For the early stop technique, following the same approach, we can get, 
    \begin{equation}
    \begin{aligned}
        \mathcal{W}_2^2 (\rp_{\delta}, \q_{T-\delta}) 
    \le K_{1}\left(e^{-2T+\delta}\left(M_2+\Tr(C)\right)+\epsilon^2T+\Tr(C)\tau^2\right).
    \end{aligned}
    \end{equation}
\end{remark}

\subsection{Result under the assumptions in Remark~\ref{rmk:relaxed app score} }\label{app:relaxed app score}
Here we discuss the relaxation of Assumption~\ref{ass:appro score lip bound} and the resulting convergence and complexity bound.
The preliminary estimate~\eqref{computation:L_2,L_3} and~\eqref{computation:lip of s+I} in Appendix~\ref{proof:w2 bound} still hold. Now we denote,
\begin{align*}
    \beta_k := \|\nabla \tilde{s}_\theta(T-t_k,\cdot)\|_\infty.
\end{align*}
Then the estimate~\eqref{computation:lip of s_theta+I} should be replaced by,
\begin{align}\label{computation:lip of s_theta+I gene}
    \forall k \in \{0,1,...,N-1\}, \quad \|\nabla \hat{s}_\theta(T-t_k,\cdot)\|_\infty\le\beta_k+L_2e^{-T+t_k}. 
\end{align}
In this section, we set the step size
\[
\tau := \sup_{k}(t_{k+1}-t_k) \le \inf_k\Big\{1,\frac{1}{(L_2+\beta_k)e}\Big\},
\]
so that
\begin{align}\label{eq:small small tau}
    \forall r>0\quad\frac{e^{r\tau}-1}{r}\le e^r\tau, \quad   
\textrm{and} \quad \forall k\in \{0,1,..,N-1\},\quad (L_2+\beta_k)e\tau \le 1.
\end{align}
Following the same approach in Appendix~\ref{proof:w2 bound}, we still have, for all $k\in\{0,1,..N-1\}$ 
\begin{align*}
    \e \left| e^{\frac{t_{k+1}}{2}}(\lx_{t_{k+1}}-\ly_{t_{k+1}})\right|^2=I_1^k+I_2^k,
\end{align*}
and $I_2^k$ satisfies~\eqref{computation:I_1^k}. With~\eqref{computation:lip of s_theta+I gene} and~\eqref{eq:small small tau}, $I_1^k$ now satisfies,
for all $f_k>0$,
\begin{align*}
    I_1^k\le& \Big(1+\big(4+3(L_2e^{-T}+\beta_ke^{-t_k})\big(e^{t_{k+1}}-e^{t_k})\Big)\e|e^{\frac{t_{k}}{2}}(\lx_{t_{k}}-\ly_{t_{k}})|^2\\
    &+(e+1)e^{T}\epsilon_k^2e(t_{k+1}-t_k).
\end{align*}
Denote,
\begin{align*}
    B_0=0 \quad \textrm{and} \quad B_k=\sum_{i=0}^{k-1}\beta_{i}(t_{i+1}-t_i), \quad k=1,2,...,N-1.
\end{align*}
Then,
\begin{align*}
    I_1^k\le& \Big(1+\big(4+3L_2)e^{-T}\big(e^{t_{k+1}}-e^{t_k})+3e(B_{k+1}-B_k)\Big)\e|e^{\frac{t_{k}}{2}}(\lx_{t_{k}}-\ly_{t_{k}})|^2\\
    &+(e+1)e^{T}\epsilon_k^2e(t_{k+1}-t_k).
\end{align*}
And the recursion for $E(k)$ becomes,
\begin{align*}
    &E(k+1)\exp\Big(-\big(4+3L_2\big)e^{-T+t_{k+1}}-3eB_{k+1}\Big)-E(k)\exp\Big(-\big(4+3L_2\big)e^{-T+t_{k}}-3eB_k\Big)\\
    \le& (e+1)e^{T}\epsilon_k^2e(t_{k+1}-t_k)+ e^3(L_0+L_2)^2\Tr(C)e^{-2T}\tau^2(e^{3t_{k+1}}-e^{3t_k}).
\end{align*}
Summing up over $k$, the bound for $\e|\lx_T-\ly_T|^2$ is of the same form as the one in Appendix~\ref{proof:w2 bound}, with the different constant coefficient,
\begin{align*}
    \e|\lx_T-\ly_T|^2\le K'_1\left(e^{-T}\e|\lx_0-\ly_0|^2+\epsilon^2T+\Tr(C)\tau^2\right),
\end{align*}
where $K'_1:=e^{4+3(L_2+eB)}\max\{(e+1)e,e^3(L_0+L_2)^2\}$.
The remaining proof follows the same approach as in Appendix~\ref{proof:w2 bound}, and ultimately yields,
\begin{align*}
     \mathcal{W}_2^2 (\rp_{0}, \q_{T}) 
    \le K'_1\left(e^{-2T}\left(M_2+\Tr(C)\right)+\epsilon^2T+\Tr(C)\tau^2\right).
\end{align*}
\subsection{Proof of Theorem~\ref{thm:MH}}\label{proof:MH}
\begin{proof}
    Recall that,
    \begin{align*}
 q_\sigma(x)&=\int_{\mathbb{R}^d}\exp(-\frac{|x-y|^2}{2\sigma^2})Q_0(dy)=\int_{B(0,R)}\exp(-\frac{|x-y|^2}{2\sigma^2})Q_0(dy),\\
 g(x)&=\log q_\sigma(x)+\frac{|x|^2}{2\sigma^2}.
    \end{align*}
    Fixing $x$, direct computation shows,
    \begin{align*}
        \nabla g(x)=&\nabla \log q_\sigma(x)+\frac{x}{\sigma^2} = \frac{\int_{B(0,R)}(\frac{y-x}{\sigma^2})\exp(-\frac{|x-y|^2}{2\sigma^2})Q_0(dy)}{\int_{B(0,R)}\exp(-\frac{|x-y|^2}{2\sigma^2})Q_0(dy)} +\frac{x}{\sigma^2} \\
        =&\frac{\int_{B(0,R)}(\frac{y}{\sigma^2})\exp(-\frac{|x-y|^2}{2\sigma^2})Q_0(dy)}{\int_{B(0,R)}\exp(-\frac{|x-y|^2}{2\sigma^2})Q_0(dy)} . 
    \end{align*}
    Taking absolute value, we get,
    \begin{equation*}
        |\nabla g(x)|=|\nabla \log q_\sigma(x)+\frac{x}{\sigma^2}|\leq \frac{R}{\sigma^2}.
    \end{equation*}
For any unit direction $z$, 
\begin{equation*}
    \begin{aligned}
        &\nabla_z \cdot \nabla_z (\log q_\sigma +\frac{|x|^2}{2\sigma^2})\\ =& \frac{1}{\sigma^2}\nabla_z\Big(\frac{\int_{B(0,R)}(y\cdot z)\exp(-\frac{|x-y|^2}{2\sigma^2})Q_0(dy)}{\int_{B(0,R)}\exp(-\frac{|x-y|^2}{2\sigma^2})Q_0(dy)}\Big) \\
        =&\frac{1}{\sigma^2}\Big(\frac{\int_{B(0,R)} (y\cdot z) (-\frac{x-y}{\sigma^2}\cdot z)\exp(-\frac{|x-y|^2}{2\sigma^2})Q_0(dy)}{ \int_{B(0,R)} \exp(-\frac{|x-y|^2}{2\sigma^2})Q_0(dy)}\\
        &-\frac{\int_{B(0,R)} (y\cdot z) \exp(-\frac{|x-y|^2}{2\sigma^2})Q_0(dy)\int_{B(0,R)} (-\frac{x-y}{\sigma^2}\cdot z)\exp(-\frac{|x-y|^2}{2\sigma^2})Q_0(dy)}{( \int_{B(0,R)} \exp(-\frac{|x-y|^2}{2\sigma^2})Q_0(dy))^2}\Big )\\
        =&\frac{1}{\sigma^4}\Big(\frac{\int_{B(0,R)} (y\cdot z)^2\exp(-\frac{|x-y|^2}{2\sigma^2})Q_0(dy)}{ \int_{B(0,R)} \exp(-\frac{|x-y|^2}{2\sigma^2})Q_0(dy)}-\frac{(\int_{B(0,R)} (y\cdot z)\exp(-\frac{|x-y|^2}{2\sigma^2})Q_0(dy))^2}{(\int_{B(0,R)} \exp(-\frac{|x-y|^2}{2\sigma^2})Q_0(dy))^2}\Big).
    \end{aligned}
\end{equation*}
    Taking absolute value again, we obtain, 
    \begin{equation*}
        |\nabla_z \cdot \nabla_zg(x)|=|\nabla_z \cdot \nabla_z(\log q_\sigma +\frac{|x|^2}{2\sigma^2})|\leq \frac{2R^2}{\sigma^4}.
    \end{equation*}
Thus,
\begin{equation*}
    \|\nabla^2g\|_\infty \le \frac{2R^2}{\sigma^4}.
\end{equation*}
\end{proof}

\subsection{Proof of Theorem~\ref{thm:w_2 bound under manifold ass}}\label{proof::w_2 bound under manifold ass}
\begin{proof}  
 The proof follows the same procedure as in~\ref{proof:w2 bound}; it suffices to substitute the corresponding values of the constants. By Corollary~\ref{cor:hession bound under manifold ass},
        \begin{equation*}
            \begin{cases}
            C=I_d,\\
            L_0=3\frac{R^2}{(1-e^{-\delta})^2},\\
            L_2=\sup_{t\ge \delta}\|e^t(I-C\A^{-1}_{t})\|=\frac{1}{1-e^{-\delta}}.\\
        \end{cases}
        \end{equation*}
        By Remark~\ref{rmk:w_2 under early stop setting} and noticing that under bounded-support assumption, $$M_2=\e|\rx_0|^2\le R^2,$$ we get,
        \begin{equation*}
        \begin{aligned}
             \mathcal{W}_2 (\rp_{\delta}, \q_{T-\delta}) \le&  e^{4+3(L_0+L_2)}e^3(L_0+L_2)^2\left(e^{-2T+\delta}\left(R^2+d\right)+\epsilon^2T+d\tau^2\right)\\
             \le& 2e^{7+4(L_0+L_2)}\left(e^{-2T+\delta}\left(R^2+d\right)+\epsilon^2T+d\tau^2\right)\\
             =&2\exp \left(7+\frac{12R^2}{(1-e^{-\delta})^2}+\frac{4}{1-e^{-\delta}}\right)\left(e^{-2T+\delta}\left(R^2+d\right)+\epsilon^2T+d\tau^2\right)
        \end{aligned}     
        \end{equation*}
        \end{proof} 
        
\subsection{Proof of Theorem~\ref{thm:bayes}}\label{proof:bayes}
\begin{proof}
    We first validate that the posterior follows Gaussian tail Assumption~\ref{ass:Gtail} with  $A=C$, $h(x)=-\frac{|G(x)-y|^2_\Sigma}{2}$,  and,
\begin{equation*}
    \begin{aligned}
        |\sqrt{C}\nabla h(x)|=&|\sqrt{C}\nabla G(x)\Sigma^{-1}(G(x)-y)|\le \|C\|^{\frac{1}{2}}(|G|_\infty+|y|)\|\Sigma\|^{-1}\|\nabla G\|_\infty,\\
        \|C\nabla^2 h(x)\|=&\|C\nabla^2 G(x)\Sigma^{-1}(G(x)-y)+C\nabla G(x)\Sigma^{-1}\nabla G(x)^T\|\\\le& \|C\|\|\Sigma\|^{-1}\Big(\|\nabla^2G\|_{\infty}(|G|_\infty+|y|)+\|\nabla G\|_\infty^2\Big).
    \end{aligned}
\end{equation*}
Then, it suffices to substitute the corresponding values of the constants in Appendix~\ref{proof:w2 bound},
\begin{align*}
    K =&  \max\{1, \|A C^{-1}\|\}=1,\\
    L_0 =& K^2 \big(\|C \nabla^2 h\|_\infty + |\sqrt{C} \nabla h|_\infty^2\big)\\
         \le &\|C\|\|\Sigma\|^{-1}\Bigg(\Big(\|\nabla^2G\|_{\infty}(|G|_\infty+|y|)+\|\nabla G\|_\infty^2\Big)+\|\Sigma\|^{-1}\|\nabla G \|^2_\infty\Big(|G|_\infty+y\Big)^2\Bigg),\\
    L_2 =& \max \{\|I-CA^{-1}\|,\|AC^{-1}-I\|\}=0. 
\end{align*}
 Now following the proof in Appendix~\ref{proof:w2 bound}, we have,   
 \begin{equation*}
     \mathcal{W}_2^2 \big(\rp_0(\cdot;y), \q_{T}(\cdot;y)\big) 
              \le K_3\Big(e^{-2T}\big(M_2+\Tr(C)\big)+\epsilon^2T+\Tr(C)\tau^2\Big),
\end{equation*}
where
\begin{equation*} 
\begin{aligned}
  K_3&=e^{4+3k_3}\max\{(e+1)e,e^3k_3^2\},\\ k_3&=\|C\|\|\Sigma\|^{-1}\Bigg(\Big(\|\nabla^2G\|_{\infty}(|G|_\infty+|y|)+\|\nabla G\|_\infty^2\Big)+\|\Sigma\|^{-1}\|\nabla G \|^2_\infty\Big(|G|_\infty+y\Big)^2\Bigg).
\end{aligned}    
\end{equation*} 
\end{proof}

\section{Theories towards the generative diffusion model in infinite dimension}\label{app:infdim}
\subsection{Motivating example towards the infinite dimensional result}\label{app:klvsw}

 We consider the following target distribution,
 \begin{align*}
     p_0(x) = \frac{1}{Z} \prod_{i=1}^d\Big(\frac{1}{2}\exp(-\frac{|x-\sqrt{C_i}|^2}{2C_i})+\frac{1}{2}\exp(-\frac{|x+\sqrt{C_i}|^2}{2C_i})\Big),
 \end{align*}
where $C$ is an $d\times d$ matrix diagonal matrix with $\{C_i\}_i$ as diagonal entries. Denote the case $d=1$ and $C_1=1$ as,\begin{align*}
    p_{0}^1(x)=\frac{1}{2\sqrt{2\pi}}
\Big(\exp (-\frac{(x-1)^2}{2})+\exp (-\frac{(x+1)^2}{2})\Big),\end{align*}
and if one considers to apply the forward process~\eqref{eq:forward process}, one can get the distribution at time $t$,
\begin{align*}
    p_{t}^1(x)=\frac{1}{2\sqrt{2\pi}}
\Big(\exp (-\frac{(x-e^{-\frac{t}{2}})^2}{2})+\exp (-\frac{(x+e^{-\frac{t}{2}})^2}{2})\Big).
\end{align*}
Simple calculation shows, 
\begin{align}\label{eq:w2vsklp0}
     \mathcal{W}_2  \big(p_0,\N(0,C) \big)\le \sqrt{2\Tr(C)} ,\quad 
    KL \big(p_0||\N(0,C) \big)=d\cdot KL \big(p_0^1||\N(0,1) \big).
\end{align}
And if one considers applying the forward process~\eqref{eq:forward process}, one can also show that,
\begin{align}\label{eq:w2vsklpt}
     \mathcal{W}_2   \big(p_t,N(0,C) \big)\le e^{-\frac{t}{2}}\sqrt{2\Tr(C)} ,\quad 
    KL \big(p_t||\N(0,C) \big)=d\cdot KL \big(p_t^1||\N(0,1) \big).
\end{align}
From~\eqref{eq:w2vsklp0} and~\eqref{eq:w2vsklpt}, we observe that \emph{when increasing the dimension of $C$ while keeping $\Tr(C)$ fixed,} the Wasserstein-$2$ bounds only scale with the trace of $C$ while the KL bounds scale with $d$.

\subsection{Defining diffusion model in infinite dimension}\label{app:infinitedim}
Now we consider a separable Hilbert space $H$ with inner product $\langle\cdot,\cdot\rangle_H$.
The forward process in $H$ has the same form as~\eqref{eq:forward process},
    \begin{equation}
     d\rx_t= -\frac{1}{2}\rx_tdt+\sqrt{C}dB_t, \quad   0\le t\le T,
     \label{eq:infinite forward process}
\end{equation}
where $C$ becomes a degenerate positive trace-class.  We still denote the marginal distributions of $\rx_t$ by $\rp_t$, which converges to the stationary distribution $\N (0, C)$ as $t \rightarrow \infty$ (\citep{da2014stochastic} Theorem 11.11). After time reversal $\lx_t=\rx_{T-t}$, $\lx_t$ satisfies the backward SDE~(\citep{pidstrigach_infinite-dimensional_2023} Theorem 1),
\begin{equation}
    d\lx_t=  \big(\frac{1}{2}\lx_t+s(T-t,\lx_t) \big)dt+\sqrt{C}d\tilde{B}_t,
    \label{eq:infinite backward process}
\end{equation}
where score function $s(t,x)$ is defined as:\begin{equation}\label{eq:infinite dimensional score}
    s(t,x):=-\frac{1}{1-e^{-t}}\e [\rx_t-e^{-\frac{t}{2}}\rx_0|\rx_t=x],
\end{equation}
which is almost surely continuous in $t$ with respect to the norm $\|\|_H$ and equal to $C\nabla \log p_t(x)$ when $H=\mathbb{R}^d$.
We consider  $s_\theta(t,x)$ to approximate the score operator in $H$ \eqref{eq:infinite dimensional score} and the scheme~\eqref{eq:explicit solver} for the sampling process.
We now introduce the infinite-dimensional Gaussian tail assumption.
\begin{assumption}\label{ass:G tail infinite dim}
    The initial data distribution $\rp_0$ has finite second moment $M_2$ and has a Gaussian tail, i.e.,
\begin{equation}\label{eq:Gaussian tail in infinite space}
    d\rp_0(x) \propto \exp \big(h(x) \big)d\N(0,A)
    (x),
\end{equation}where 
$A$ is a degenerate positive trace-class operator that is simultaneously diagonalizable with 
$C$, $A$ and $C$ correspond to the same Cameron–Martin space, and both $AC^{-1}$ and $A^{-1}C$ are bounded linear maps. The function  $h$ is two times differentiable and satisfies, \begin{equation}
\begin{aligned}
      &\|\sqrt{C}\nabla h\|_{H,\infty}:=\sup_x \|\nabla \sqrt{C}h(x)\|_{H} < \infty, \\
        &\|C\nabla^2 h\|_{\mathcal{L}(H),\infty}:=\sup_x \|C\nabla^2 h(x)\|_{\mathcal{L}(H)} < \infty .
\end{aligned}
\end{equation} 
\end{assumption}

To extend our Corollary~\ref{cor:score lip exp decay bound} and Theorem~\ref{thm:w2 bound} to the infinite-dimensional case, we first follow the approach in Appendix E of \citep{pidstrigach2023infinite} that projects $H$ onto a finite-dimensional subspace $H^D$, and  approximates the infinite-dimensional case via the results established on $H^D$.

Suppose $C$ ($A$, correspondingly) has an orthonormal basis $e_i$ of eigenvectors
and corresponding non-negative eigenvalues $c_i \ge 0$ ($a_i \ge 0$, correspondingly), we define the linear span of the first $D$ eigenvectors as\begin{equation*}
    H^D=\operatorname{span} \{e_1,e_2,\ldots , e_D\}.
\end{equation*}
Let $P^D: H \rightarrow H^D$ be the orthogonal projection onto $H^D$.
We define the finite-dimensional approximations of $\mu_{data}$ by $\mu_{data}^D={P^D}_\#(\mu_{data})$ and discretize the forward 
process by $\rxdt=P^D\rx_t$, then $(\rxdt)_{t\ge0}$ will satisfy,
\begin{align}
     d\rxdt= -\frac{1}{2}\rxdt dt+\sqrt{P^DCP^D}dB_t.
\end{align} We denote the marginal distribution of $\rxdt$ by $\rpdt$, and its density by $p^D_t$.
The corresponding backward process will be,
\begin{equation}
     d\lxdt= (\frac{1}{2}\lxdt+s^D(T-t,\lxdt))dt+\sqrt{P^DCP^D}d\tilde{B}_t,
\end{equation}
where $s^D(t,x)=P^D\e[s(t,\rx_t)|P^D\rx_t=x]=C^D\nabla\log p_t^D(x)$.

Then we show the projected distribution still follows the Gaussian tail assumption in finite dimension, summarize as follow.
\begin{lemma}
    The projected initial distribution $\rpdt$ satisfies the finite-dimensional Gaussian tail Assumption~\ref{ass:Gtail},
    \begin{align*}
        d\rpdo (x^D) \propto \exp \big(h^D(x^D) \big)d\N(0,A^D)(x^D),
    \end{align*}
    where 
    \begin{equation}\label{eq:finite dimensional tail bounded by infinite dimensional tail}
        \begin{aligned}
         &|\sqrt{C^D}\nabla h^D|_\infty\le \|\sqrt{C}\nabla h\|_{H,\infty},\\&\|C^D\nabla^2h^D\|_\infty\le\|C\nabla^2h\|_{\mathcal{L}(H),\infty}^2+\|\sqrt{C}\nabla h\|_{H,\infty}^2.
    \end{aligned}
    \end{equation}    
\end{lemma}
\begin{proof}
The term $\exp (h^D(x^D))$ is given by,
\begin{equation*}
\begin{aligned}
     \exp(h^D(x^D))&=\int \exp \big(h(x^D,x^{D+1:\infty}) \big)d\mathcal{N}(0,A^{D+1:\infty})(x^{D+1:\infty})\\
     &=\e_{\mathcal{N}(0,A)}[\exp \big(h(X) \big)|X^D=x^D].
\end{aligned}   
\end{equation*}
Then, the gradient is given by,
\begin{equation*}
\begin{aligned}
    \nabla h^D(x^D)&=\frac{\int \nabla _{x^D} h(x^D,x^{D+1:\infty})\exp \big(h(x^D,x^{D+1:\infty}) \big)d\mathcal{N}(0,A^{D+1:\infty})(x^{D+1:\infty})}{\int \exp \big(h(x^D,x^{D+1:\infty}) \big)d\mathcal{N}(0,A^{D+1:\infty})(x^{D+1:\infty})}\\
    &=\e_{\tilde{P}_0}[\nabla_{x^D}h(X)|X^D=x^D],
\end{aligned}
\end{equation*}
where $d\tilde{P}_0(x)\propto \exp \big(h(x)\big)d\N (0,A),$
and the Hessian is given by,
\begin{equation*}
    \begin{aligned}
        \nabla^2h^D(x^D)=&\frac{\int \nabla^2 _{x^D} h(x^D,x^{D+1:\infty})\exp \big(h(x^D,x^{D+1:\infty}) \big)d\mathcal{N}(0,A^{D+1:\infty})(x^{D+1:\infty})}{\int \exp \big(h(x^D,x^{D+1:\infty}) \big)d\mathcal{N}(0,A^{D+1:\infty})(x^{D+1:\infty})}\\
        &-\Big(\frac{\int \nabla _{x^D} h(x^D,x^{D+1:\infty})\exp \big(h(x^D,x^{D+1:\infty}) \big)d\mathcal{N}(0,A^{D+1:\infty})(x^{D+1:\infty})}{\int \exp \big(h(x^D,x^{D+1:\infty}) \big)d\mathcal{N}(0,A^{D+1:\infty})(x^{D+1:\infty})}\Big)^{\otimes2}\\
        =&\e_{\tilde{P}_0}[\nabla^2_{x^D}h(X)|X^D=x^D]-(\e_{\tilde{P}_0}[\nabla_{x^D}h(X)|X^D=x^D])^{\otimes 2}.
    \end{aligned}
\end{equation*}
Taking the absolute value, we get,
\begin{equation*}
\begin{aligned}
     &|\sqrt{C^D}\nabla h^D(x^D)|\le |\sqrt{C^D}\nabla_{x^D} h|_\infty\le \|\sqrt{C}\nabla h\|_{H,\infty},\\
     &\|C^D\nabla^2h^D(x^D)\|\le \|C^D\nabla_{x^D}^2h\|_{\infty}+|\sqrt{C^D}\nabla _{x^D}h|_{\infty}^2\le\|C\nabla^2h\|_{\mathcal{L}(H),\infty}^2+\|\sqrt{c}\nabla h\|_{H,\infty}^2.  
\end{aligned}
\end{equation*}
Thus,
\begin{equation*}
    \begin{aligned}
         &|\sqrt{C^D}\nabla h^D|_\infty\le \|\sqrt{C}\nabla h\|_{H,\infty},\\&\|C^D\nabla^2h^D\|_\infty\le\|C\nabla^2h\|_{\mathcal{L}(H),\infty}^2+\|\sqrt{C}\nabla h\|_{H,\infty}^2. 
    \end{aligned}
\end{equation*}
\end{proof}
Define, \begin{equation*}
        \begin{aligned}
            &\A_t:=Ae^{-t}+C(1-e^{-t}),\\
            &\A^D_t:=A^De^{-t}+C^D(1-e^{-t}),
        \end{aligned}
    \end{equation*} 
 and 
    \begin{equation}\label{eq:infinite dimension modi score}
        \begin{aligned}    &\tilde{s}^D(t,x):=s^D(t,x)+C^D(\A^D_t)^{-1}x,\\
            &\tilde{s}(t,x):=s(t,x)+C\A_t^{-1}x.
        \end{aligned}
    \end{equation}
Since the projected initial distribution $\rpdo$ satisfies the finite-dimensional Gaussian tail assumption (Assumption~\ref{ass:Gtail}),  Corollary~\ref{cor:score lip exp decay bound} implies that there exist constants $L_0^D ,L^D_1\ge 0$ such that,
    \begin{equation*}
         \|\nabla \tilde{s}^D(t,\cdot)\|_{\mathcal{L}(H^D),\infty} \le L^D_0e^{-t},\ \quad
        \|\tilde{s}^D(t,\cdot)\|_{H^D,\infty} \le L_1^De^{-\frac{t}{2}}, \quad \forall t \ge 0.
    \end{equation*}
Here the constant $L_0^D$ and $L_1^D$ is given by,
\begin{equation*}
\begin{cases}
    K^D =  \max\{1, \|A^D (C^D)^{-1}\|\}\le \max\{1, \|A C^{-1}\|\}:=K,\\
    L^D_0 = (K^D)^2 \big(\|C^D \nabla^2 h^D\|_\infty + |\sqrt{C^D} \nabla h^D|_\infty^2\big)\le K^2 \big(\|C \nabla^2 h\|_{\mathcal{L}(H),\infty} + 2\|\sqrt{C} \nabla h\|_{H,\infty}^2\big),\\
    L_1^D = K^D \|C^D\|^{1/2} |\sqrt{C^D} \nabla h^D|_\infty\le K\|C\|^{\frac{1}{2}}\|\sqrt{C}\nabla h\|_{H,\infty}.
\end{cases}
\end{equation*}
    By \citep{pidstrigach2023infinite} lemma 5 (listed as Proposition~\ref{prop:L_2 and a.s convergence}), we have 
    \begin{equation}\label{eq:infinite mod score}
        \tilde{s}^D(t,\rxdt) \rightarrow \tilde{s}(t,\rx_t)=s(t,\rx_t)+C\A_t^{-1}\rx_t, \quad \text{both}\ a.s. \text{ and }\ L_2,\quad \text{as} \ D \rightarrow \infty.
    \end{equation}
 Thus, we obtain the following regularity estimation for the modified score,
 \begin{theorem} \label{thm:infinte tail app}
     Under the infinite dimensional Gaussian tail Assumption~\ref{ass:G tail infinite dim}, the modified score $\tilde{s}$ defined by~\eqref{eq:infinite dimensional score} and~\eqref{eq:infinite dimension modi score} satisfies,
\begin{equation}
     \label{eq:infinte tail app}
        \|\nabla \tilde{s}(t,\cdot)\|_{\mathcal{L}(H),\infty}\le \tilde{L}_0e^{-t},\quad
        \|\tilde{s}(t,x)\|_{H,\infty} \le \tilde{L}_1e^{-\frac{t}{2}}, \quad \forall  t>0.
 \end{equation}
   where 
\begin{equation*}
\begin{cases}
    K =  \max\{1, \|A C^{-1}\|\},\\
    \tilde{L}_0 =  K^2 \big(\|C \nabla^2 h\|_{\mathcal{L}(H),\infty} + 2\|\sqrt{C} \nabla h\|_{H,\infty}^2\big),\\
    \tilde{L}_1 = K\|C\|^{\frac{1}{2}}\|\sqrt{C}\nabla h\|_{H,\infty}.
\end{cases}
\end{equation*}
 \end{theorem}
\eqref{eq:infinte tail app} pertains to Corollary~\ref{cor:score lip exp decay bound} formulated in the infinite-dimensional case. To derive an error bound, we also need Assumption~\ref{ass:error} and \ref{ass:appro score lip bound} under an infinite-dimensional setting, listed as follows.
\begin{assumption}\label{ass:infinite error}
    For each time discretization point $t_k$, $ 0 \le k \le N-1$, the approximated score $\tilde{s}_\theta$ satisfies,
    \begin{equation*}
    \begin{aligned}
        &\|\nabla \tilde{s}_\theta(T-t_k,\cdot)\|_{\mathcal{L}(H),\infty} \le \tilde{L}_0e^{-T+t_k},\\        
        &\frac{1}{T}\sum_{k=0}^{N-1}(t_{k+1}-t_k)\e\|s(T-t_k,\rx_{T-t_k})-s_\theta(T-t_k,\rx_{T-t_k})\|_H^2 \le \epsilon^2.
    \end{aligned}    
    \end{equation*}
\end{assumption}
To measure the distance between the samples and the target distribution, we introduce the Wasserstein-2-distance on the Hilbert space $H$,
\[ \mathcal{W}_2 (\mu, \nu) = \left(\inf_{\pi} \int \| x - y
\|_H^2 d \pi (x, y)\right)^{1/2},  \]
where $\pi$ runs over all measures on $H\times H$ with marginals $\mu$ and
$\nu$.

Now we present the convergence bound aligned with the main  Theorem~\ref{thm:w2 bound} under the infinite-dimensional setting for completeness.  
\begin{theorem}\label{thm:inf-dim-w2-bound}
    Suppose Assumptions \ref{ass:G tail infinite dim} and \ref{ass:infinite error} hold under the infinite infinite-dimensional setting. Then, sampling via scheme~\eqref{eq:explicit solver} with sufficiently small step size $\tau:=\sup_{k}(t_{k+1}-t_k)$ yields
    \begin{equation}
       \mathcal{W}_2^2 (\rp_{0}, \q_{T}) 
    \le \tilde{K}_1\Big(e^{-2T}\left(M_2+\Tr(C)\right)+\epsilon^2T+\Tr(C)\tau^2\Big),
    \end{equation}
    where $\tilde{K}_1=e^{4+3(\tilde{L}_0+L_2)}\max\{(e+1)e,e^3(\tilde{L}_0+L_2)^2\}$. 
\end{theorem}
\begin{proof}
First, we list the preliminary estimates as in Appendix~\ref{proof:w2 bound}. Denote
\begin{align*}
    \hat{s}(t,x):=s(t,x)+x,\quad \hat{s}_\theta(t,x):=s_\theta(t,x)+x
\end{align*}
and 
\begin{align*}
    L_2:=\max \{\|I-CA^{-1}\|,\|AC^{-1}-I\|\}=\sup_{t}e^t\|I-C\bar{A}_t^{-1}\|.
\end{align*}
By Theorem~\ref{thm:infinte tail app} and Assumption~\ref{ass:infinite error}, we have,
\begin{align*}
    &\forall t\ge 0,\quad \|\nabla \hat{s}(t,\cdot)\|_{\mathcal{L}(H),\infty}\le (\tilde{L}_0+L_2)e^{-t},\\
     &\forall k\in \{0,1,...,N-1\},\quad \|\nabla \hat{s}_\theta(T-t_k,\cdot)\|_{\mathcal{L}(H),\infty}\le (\tilde{L}_0+L_2)e^{-T+t_k}.
\end{align*}
We set the step size
\[
\tau := \sup_{k}(t_{k+1}-t_k) \le \min\Big\{1,\frac{1}{(L_0+L_2)e}\Big\},
\]
so that
\begin{align*}
    \forall r>0,\quad\frac{e^{r\tau}-1}{r}\le e^r\tau ,\quad  
\textrm{and} \quad (L_0+L_2)e\tau \le 1.
\end{align*}
The proof proceeds in the same manner as in Appendix~\ref{proof:w2 bound}. However, Lemma~\ref{lem:rever martingale} is not available here, since in the infinite-dimensional setting there is no probability density in the sense of Lebesgue measure. Therefore, we make appropriate adjustments at the points where Lemma~\ref{lem:rever martingale} would otherwise be used. By Theorem~\ref{thm:infinte tail app}, $\hat{s}(T-t,\lx_t)$ satisfies,
\begin{align*}
    \e\|\hat{s}(T-t,\lx_t)\|_H^2=&\e\|\tilde{s}(T-t,\lx_t)+(I-C\A_{T-t}^{-1})\lx_t\|^2\\
    \le& 2\e \|\tilde{s}(T-t,\lx_t)\|_H^2+2\e\|(I-C\A_{T-t}^{-1})\lx_t\|^2\\
    \le&2\tilde{L}_1^2e^{-(T-t)}+2L_2^2e^{-2(T-t)}\e\|\lx_t\|^2  \\
    \le& 2\tilde{L}_1^2e^{-(T-t)}+2L_2^2e^{-2(T-t)}\max \{M_2,\Tr(C)\}\\
    \le& \infty.
\end{align*}
Thus, by Proposition~\ref{prop:L_2 and a.s convergence}, $\hat{s}^D(T-t,\lxdt):=s^D(T-t,\lxdt)$ converge to $\hat{s}(T-t,\lx_t)$ in $L^2$. By Lemma~\ref{lem:rever martingale}, $e^{-\frac{t}{2}}\hat{s}^D(T-t,\lxdt)$ is a continuous time martingale. Thus, by Doob's $L^2-$inequality, for any $D,N \in \N^{+}$
\begin{align*}
    \e [\sup_{0\le t \le T}|e^{-\frac{t}{2}}\hat{s}^D(T-t,\lxdt)-e^{-\frac{t}{2}}\hat{s}^N(T-t,\lxnt)|^2]\le 4e^{-T}\e|\hat{s}^D(0,\lxdT)-\hat{s}^N(0,\lxnT)|^2.
\end{align*}
Since the right side is Cauchy, the left side is also Cauchy, i.e., continuous time martingales $e^{-\frac{t}{2}}\{s^D(T-t,\lxdt)\}_D$ form a Cauchy sequence under the norm,
\begin{align*}
    \|s\|=\e[\sup_{t\in [0,T]}|s_t|^2].
\end{align*}
The continuous martingales are closed with respect to this norm~(\citep{karatzas2014brownian} Section 1.3), so $e^{-\frac{t}{2}}s(T-t,\lx_t)$ is also a continuous time martingale.

Now we follow the approach in Appendix~\ref{proof:w2 bound}. Coupling $\lx_t$ and $\ly_t$ with the same Brownian motion, and using the fact that $e^{-\frac{t}{2}}s(T-t,\lx_t)$ is a martingale, we obtain,
\begin{align*}
     \e \left| e^{\frac{t_{k+1}}{2}}(\lx_{t_{k+1}}-\ly_{t_{k+1}})\right|^2=\tilde{I}_1^k+\tilde{I}_2^k,
\end{align*}
where 
\begin{align*}
    \tilde{I}_1^k=&\e\Bigg| e^{\frac{t_{k}}{2}}(\lx_{t_{k}}-\ly_{t_{k}})+\int _{t_k}^{t_{k+1}}e^{t-\frac{t_k}{2}}\left(\hat{s}_\theta(T-t_k,\lx_{t_k})-\hat{s}_\theta(T-t_k,\ly_{t_k})\right)dt\\
    &\quad+\int _{t_k}^{t_{k+1}}e^{t-\frac{t_k}{2}}\left(\hat{s}(T-t_{k},\lx_{t_k})-\hat{s}_\theta(T-t_k,\lx_{t_k})\right)dt\Bigg|^2,
\end{align*}
and 
\begin{align*}
    \tilde{I}_2^k=\e \Bigg |\int_{t_k}^{t_{k+1}}e^t\left(e^{-\frac{t}{2}}\hat{s}(T-t,\lx_t)-e^{-\frac{t_k}{2}}\hat{s}(T-t_k,\lx_{t_k})\right)dt\Bigg|^2.
\end{align*}
For $e^{-\frac{t}{2}}\{s^D(T-t,\lxdt)\}_D$, using the estimate for $I_2^k$ in Appendix~\ref{proof:w2 bound}, we have,
\begin{align*}
    &\e \Bigg |\int_{t_k}^{t_{k+1}}e^t\left(e^{-\frac{t}{2}}\hat{s}^D(T-t,\lxdt)-e^{-\frac{t_k}{2}}\hat{s}(T-t_k,\overset{\leftarrow}{X^D_{t_k}})\right)dt\Bigg|^2\\
    \le & e^3(\tilde{L}_0+L_2)^2\Tr(C)e^{-2T+3t_k}\tau(t_{k+1}-t_k).
\end{align*}
Since the $L^2$ convergence of $e^{-\frac{t}{2}}\hat{s}^D(T-t,\lxdt)$ to $e^{-\frac{t}{2}}\hat{s}(T-t,\lx_t)$ is uniform in time, we obtain,
\begin{align*}
    \tilde{I}^2_k \le e^3(\tilde{L}_0+L_2)^2\Tr(C)e^{-2T+3t_k}\tau(t_{k+1}-t_k).
\end{align*}

 At this stage, we have modified the parts of proof in Appendix~\ref{proof:w2 bound} that rely on Lemma~\ref{lem:rever martingale} and obtained analogous results. The remaining steps follow directly from Appendix~\ref{proof:w2 bound}, where the corresponding formulas naturally extend to the infinite-dimensional setting.
\end{proof}
Here, we list the lemma used to extend our analysis to infinite dimensions.
\begin{proposition}{(\cite{pidstrigach2023infinite}, Lemma 5)}\label{prop:L_2 and a.s convergence}
    Let $H$ be a separable Hilbert space and $Z, \tilde{Z}$ be two random variables taking values in $H$. Let $e_i$ be an orthonormal basis of $H$. Denote by $H^D = \operatorname{span}\{e_1,e_2,\ldots, e_D\}$ and by $P^D$ the projection onto $H^D$. Furthermore, let $Z^D=P^D\e[Z|P^D\tilde{Z}]$. If  $\e\|\e[Z|\tilde{Z}]\|^2_H < \infty$, then $Z^D \rightarrow \e[Z|\tilde{Z}]$ in $ L^2$ and almost surely.
\end{proposition}

\section{Table of constants}\label{app:constants}
\begin{table}[h]
\caption{Explicit expressions for the constants in the derivations}
\label{tab:constants}

\begin{center}
\scriptsize
\renewcommand{\arraystretch}{1.5}
\begin{tabular}{@{}p{3.6cm}p{12.0cm}@{}}
\toprule
Constant  &  Expression\\
\midrule
$K$ &  $\max \{1,\|AC^{-1}\|,\|A^{-1}C\|\}$ \\
$L_0$ &  $K^2(\|C\nabla^2h\|_\infty+|\sqrt{C}\nabla 
h|_\infty^2)$ \\
$\tilde{L}_0$ & $ K^2 \big(\|C \nabla^2 h\|_{\mathcal{L}(H),\infty} + 2\|\sqrt{C} \nabla h\|_{H,\infty}^2\big)$\\
$L_1$ &  $K\|C\|^{\frac{1}{2}}|\sqrt{C}\nabla h|_\infty$\\
$\tilde{L}_1$ & $K\|C\|^{\frac{1}{2}}\|\sqrt{C}\nabla h\|_{H,\infty}$\\
$L_2$ &  $\max \{\|I-CA^{-1}\|,\|AC^{-1}-I\|\}$\\
$M_0$ &  $\max \{\e|\rx_0|^2,\Tr{C},1\}$\\
$M_2$ &  $\e|\rx_0|^2$ \\
$K_1$ &   $e^{4+3(L_0+L_2)}\max\{(e+1)e,e^3(L_0+L_2)^2\}$\\
$K'_1$ & $e^{4+3(L_2+eB)}\max\{(e+1)e,e^3(L_0+L_2)^2\}$ \\
$\tilde{K}_1$ & $e^{4+3(\tilde{L}_0+L_2)}\max\{(e+1)e,e^3(\tilde{L}_0+L_2)^2\}$\\
$K_2$ & $2\exp \left(7+\frac{12R^2}{(1-e^{-\delta})^2}+\frac{4}{1-e^{-\delta}}\right)$\\
$k_3$ & $\|C\|\|\Sigma\|^{-1}\Bigg(\Big(\|\nabla^2G\|_{\infty}(|G|_\infty+|y|)+\|\nabla G\|_\infty^2\Big)+\|\Sigma\|^{-1}\|\nabla G \|^2_\infty\Big(|G|_\infty+y\Big)^2\Bigg)$\\
$K_3$ & $e^{4+3k_3}\max\{(e+1)e,e^3k_3^2\}$\\
\bottomrule
\end{tabular}
\end{center}
\end{table}

\end{document}